\documentclass[10pt,twocolumn,letterpaper]{article}

\usepackage{cvpr}              

\usepackage{graphicx}
\usepackage{amsmath}
\usepackage{amssymb}
\usepackage{booktabs}
\usepackage{csquotes}
\usepackage{algorithm}
\usepackage{algorithmicx}
\usepackage{algpseudocode}
\usepackage{multirow}
\usepackage{animate}

\usepackage[pagebackref,breaklinks,colorlinks]{hyperref}

\usepackage[normalem]{ulem}
\usepackage[capitalize]{cleveref}
\crefname{section}{Sec.}{Secs.}
\Crefname{section}{Section}{Sections}
\Crefname{table}{Table}{Tables}
\crefname{table}{Tab.}{Tabs.}

\def\cvprPaperID{1683} 
\def\confName{CVPR}
\def\confYear{2022}

\begin{document}

\title{Exploring Depth Contribution for Camouflaged Object Detection}

\author{
Mochu Xiang$^{1}$~
Jing Zhang$^{2}$~
Yunqiu Lv$^{1}$~
Aixuan Li$^{1}$~
Yiran Zhong$^{3}$~ 
Yuchao Dai$^{1,\star}$\\
$^1$ Northwestern Polytechnical University\quad
$^2$ Australian National University\quad
$^3$ SenseTime\\
}
\maketitle

\begin{abstract}
Camouflaged object detection (COD) aims to segment camouflaged objects hiding in the environment, which is challenging due to the similar appearance of camouflaged objects and their surroundings. Research in biology suggests depth can provide useful object localization cues for camouflaged object discovery. In this paper, we study the depth contribution for camouflaged object detection, where the depth maps are generated
with existing monocular depth estimation (MDE) methods. Due to the domain gap between the MDE dataset and our COD dataset, the generated depth maps are not accurate enough to be directly used. We then introduce two solutions to avoid the noisy depth maps from dominating the training process. Firstly, we present an auxiliary depth estimation branch (\enquote{ADE}), aiming to regress the depth maps. We find that \enquote{ADE} is especially necessary for our \enquote{generated depth} scenario. Secondly, we introduce a multi-modal confidence-aware loss function via a generative adversarial network to weigh the contribution of depth for camouflaged object detection.
Our extensive experiments on various camouflaged object detection datasets explain that the existing \enquote{sensor depth} based RGB-D segmentation techniques work poorly with \enquote{generated depth}, and our proposed two solutions work cooperatively, achieving effective depth contribution exploration for
camouflaged object detection.
\end{abstract}

\section{Introduction}
As a key example of evolution by natural selection, camouflage is widely adopted by the preys in the wild to
reduce their possibility of being detected
by their predators \cite{copeland1997models}. Camouflaged object detection (COD) is the technique that segments the whole scope of the camouflaged object.
It has wide applications in a variety of fields, such as military (\eg military camouflaged pattern design \cite{hall2020platform}), agriculture (\eg pest identification \cite{lev2004plant}), medicine (\eg polyp segmentation \cite{fan2020pranet}) and ecological protection (\eg wildlife protection \cite{nafus2015hiding,wilson2019experimental}. Due to both scientific value and application value, COD deserves well exploration.




Compared with the generic object detection \cite{redmon2016you,he2017mask} or segmentation techniques \cite{chen2017deeplab,ronneberger2015u}, COD is more challenging as the foreground objects usually share a very similar appearance to their surroundings. The visual cues for object identification, \eg texture, contrast, edge, color, and object size, are vulnerable to attack from the basic camouflage strategies, \eg background matching and disruptive coloration \cite{thayer1918concealing, price2019background}.
Although some recent deep learning based studies \cite{le2019anabranch, yan2020mirrornet, fan2020camouflaged, dong2021towards} have shown favorable performance, the misleading information in camouflage hinders the network from learning the most discriminative features of camouflage. We argue that more visual perceptual knowledge about camouflaged objects can be beneficial.

Research in biology suggests that depth provides 3D geometric information that makes the observer more sensitive to the true boundary and thus enables camouflage less effective \cite{adams2019disruptive,kelman2008review}. Accordingly, combining depth with RGB images can serve as a new way to solve the challenges in camouflaged object detection.
Further, the visual system of many species operates in the real 3D scenes, and a lot of recent works for object segmentation \cite{hu2019acnet,fan2020bbsnet, ucnet_sal,qian2020bi} have integrated depth map as another modal on top of RGB images to extract features about the 3D layout of the scene and object shape information.
As far as we know, there exist no deep camouflaged object detection models exploring the depth contribution.
In this paper, we present the first depth-guided camouflaged object detection network to study the contribution of depth for the COD task.

\begin{figure*}[!ht]
\begin{tabular}{c@{ } c@{ } c@{ } c@{ }c@{ } c@{ } c@{ } c@{ }}
\centering
   {\includegraphics[width=0.119\linewidth]{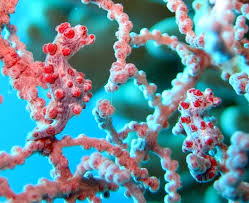}}&
    {\includegraphics[width=0.119\linewidth]{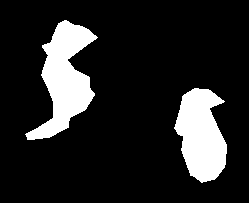}}&
    {\includegraphics[width=0.119\linewidth]{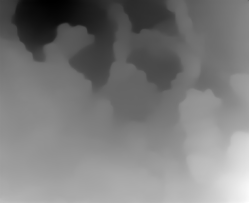}}&
    {\includegraphics[width=0.119\linewidth]{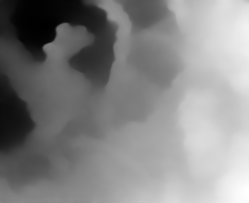}}&
    {\includegraphics[width=0.119\linewidth]{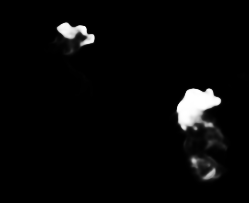}}&
    {\includegraphics[width=0.119\linewidth]{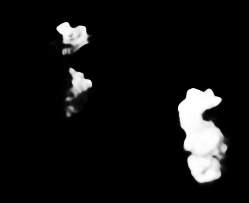}}&
    {\includegraphics[width=0.119\linewidth]{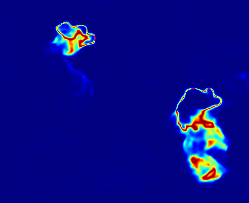}}&
    {\includegraphics[width=0.119\linewidth]{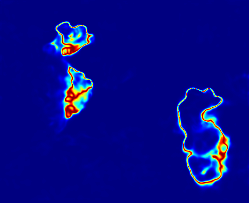}}\\
    {\includegraphics[width=0.119\linewidth]{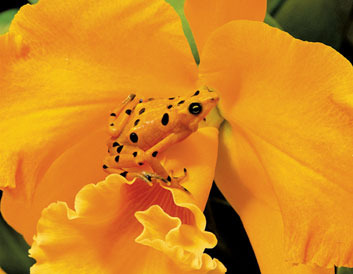}}&
    {\includegraphics[width=0.119\linewidth]{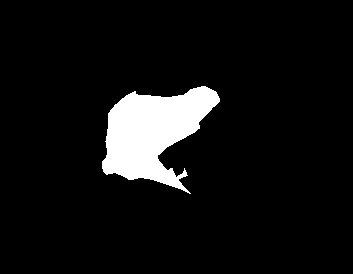}}&
    {\includegraphics[width=0.119\linewidth]{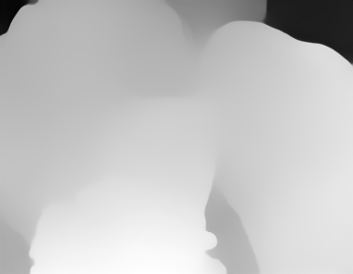}}&
    {\includegraphics[width=0.119\linewidth]{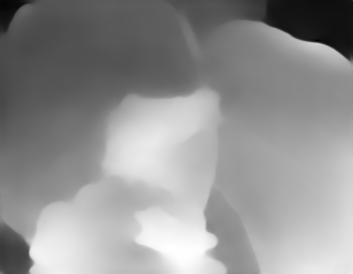}}&
    {\includegraphics[width=0.119\linewidth]{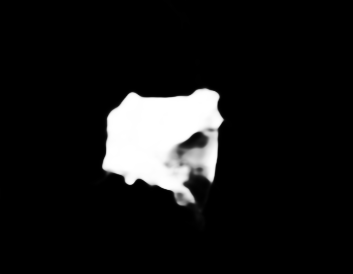}}&
    {\includegraphics[width=0.119\linewidth]{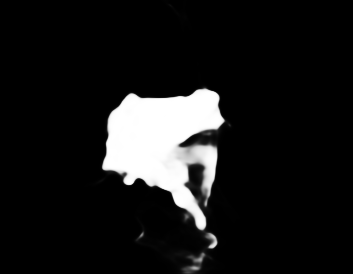}}&
    {\includegraphics[width=0.119\linewidth]{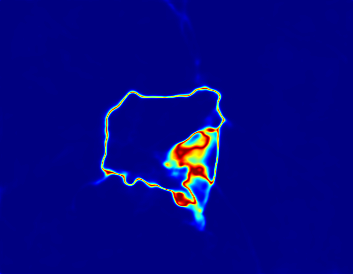}}&
    {\includegraphics[width=0.119\linewidth]{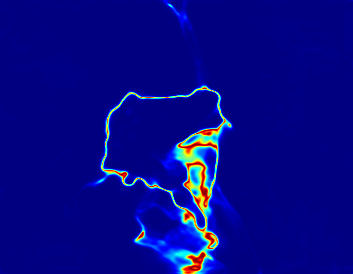}}\\
    \footnotesize{Image} &
     \footnotesize{GT} &\footnotesize{Depth} &\footnotesize{$d'$}&
    \footnotesize{Ours\_{RGB}} &
     \footnotesize{Ours\_{RGBD}} &\footnotesize{$U_{\text{rgb}}$} &\footnotesize{$U_{\text{rgbd}}$} \\
   \end{tabular}
    \caption{The auxiliary depth estimation branch (\enquote{ADE}) (for depth regression ($d'$)) and the generative adversarial network \cite{gan_raw} (for uncertainty estimation ($U_{rgb}$ and $U_{rgbd}$)) work cooperatively to produce effective camouflage maps (Ours\_{RGB} and Ours\_{RGBD}). 
    }
    \label{fig:our_solution_outputs}
\end{figure*}

As there exists no RGB-D camouflaged object detection dataset, we generate depth map of the COD training dataset \cite{le2019anabranch,fan2020camouflaged} with existing monocular depth estimation method \cite{MiDaS_Ranftl_2020_TPAMI}. With the generated depth, a straightforward solution to achieve RGB-D COD is through multi-modal fusion strategies \cite{fan2020bbsnet,dmra_iccv2019}.
However, the conventional monocular depth estimation models are trained on natural images, where there may not exist any camouflaged objects. The domain gap between the monocular depth estimation training dataset and COD training dataset leads to less accurate (or noisy) depth maps as shown in Fig.~\ref{fig:our_solution_outputs} \enquote{Depth}. Directly training with the noisy depth map may lead to an over-fitting model that generalizes poorly on the testing dataset \cite{on_calibration}.

To effectively use the depth information, we propose two main solutions, namely auxiliary depth estimation branch (\enquote{ADE}) and a multi-modal confidence-aware loss function via a generative adversarial network (GAN) \cite{gan_raw}.
The former aims to regress the depth maps supervised by the generated depth maps from existing MDE method, which is proven
especially necessary for our \enquote{generated depth} scenario. With the latter, the stochastic attribute of the network makes it possible to estimate the confidence of model prediction, which will serve as the weight for our multi-modal loss function. The basic assumption is that compared with a less confident prediction, a highly confident prediction should contribute more to model updating. We show the estimated uncertainty maps $U_{\text{rgb}}$ and $U_{\text{rgbd}}$ (the inverse confidence map) of each modal (RGB and RGB-D in this paper) in Fig.~\ref{fig:our_solution_outputs}, which validate the multi-modal confidence-aware learning, leading to better camouflage maps from the RGB-D branch (\enquote{Ours\_RGBD}).
Our main contributions can be summarized as: \textbf{1}) We advocate the contribution of depth for camouflaged object detection, and propose the first depth-guided camouflaged object detection network; \textbf{2}) We introduce an auxiliary depth estimation branch, where the RGB-D feature fusion is achieved through multi-task learning instead of multi-modal fusion within a single task, which is proven especially effective in our \enquote{generated depth} scenario; \textbf{3}) We present multi-modal confidence-aware loss function via generative adversarial network \cite{gan_raw} to effectively weigh the contribution of depth for model updating.

\section{Related Work}
\noindent\textbf{Camouflaged Object Detection}
Camouflage is a defense mechanism for animals to change their salient signatures and become invisible in the environment \cite{copeland1997models}. In early time, researchers have developed extensive methods using handcrafted features, \eg~edge, brightness, color, gradient, texture, to detect the camouflaged object \cite{tankus2001convexity,bhajantri2006camouflage,feng2015camouflage, xue2016camouflage,pike2018quantifying, fusion_camo}. However, these algorithms are far from  practical applications because the well-performed camouflage is skilled in breaking the low-level features. Recent research in COD \cite{zheng2018detection, fang2019camouflage,le2019anabranch, yan2020mirrornet, fan2020camouflaged, dong2021towards,Yang_2021_ICCV_COD_Uncertainty} mainly focus on using deep neural networks to
extract high-level semantic features to discriminate the concealed object from the complex scenarios. In addition, \cite{kelman2008review} claims that cuttlefish makes use of low-level cues to disrupt the perception of visual depth and therefore disguise their 3D shape. As described in \cite{penacchio2015three}, seeing the environment in 3D makes the presence of the object easier to be discerned. \cite{adams2019disruptive} verifies that the real 3D structure information provided by the depth is advantageous to overcome the disruption caused by edge enhancement. 
Based on these discussions, we argue that a better understanding of the 3D layout of the scene can be beneficial for more effective camouflaged object detection.
In this paper, we aim to explore depth contribution for camouflaged object detection.





\begin{figure*}[thp]
   \begin{center}
   \begin{tabular}{c@{ }}
   {\includegraphics[width=0.95\linewidth]{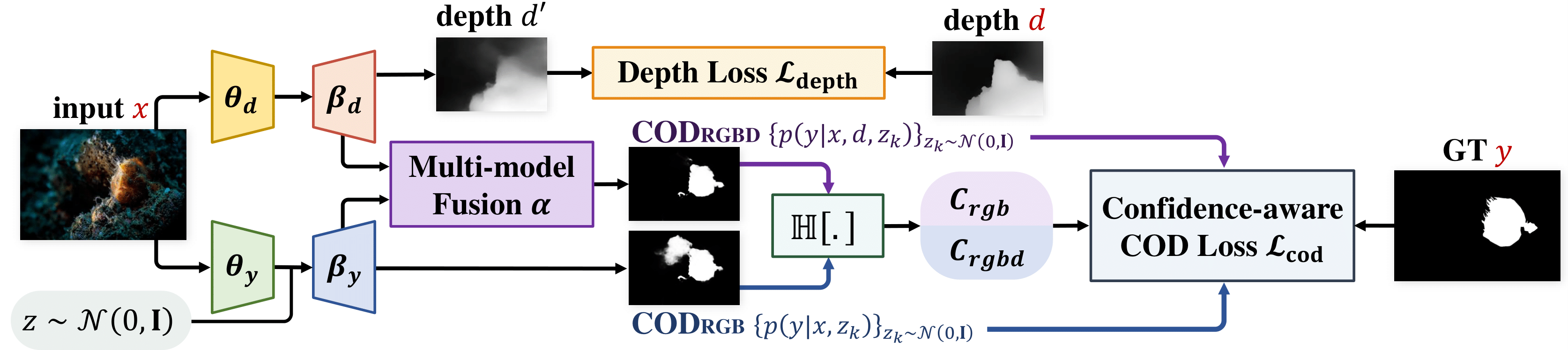}}\\
   \end{tabular}
   \end{center}
    \caption{
    Our depth contribution exploration network takes RGB image $x$ as input to achieve three main tasks: 1) RGB image based camouflaged object detection; 2) auxiliary depth estimation and 3) RGB-D camouflaged object detection.
    In addition, the proposed multi-modal confidence-aware loss function explicitly evaluates depth contribution with confidence of prediction ($C_{\text{rgb}}$ and $C_{\text{rgbd}}$) as indicator, where the confidence is obtained with the proposed probabilistic model
    via generative adversarial network \cite{gan_raw}. We only show the camouflage generator and the discriminator is introduced in Section \ref{sec:depth_contribution_exploration}.
    }
    \label{fig:network_overview}
\end{figure*}

\noindent\textbf{Depth-Guided Segmentation}
For segmentation tasks, many early work has shown that depth information can reduce the ambiguity of RGB features in complex scenarios \cite{silberman2011indoor, ren2012rgb,gupta2013perceptual,deng2015semantic,peng2014rgbd, ren2015exploiting, feng2016local, wang2017rgb}.
In recent years, the deep multi-modal fusion models
for segmentation have achieved significant performance improvement. \cite{farabet2012learning} proposes a simple early fusion strategy by concatenating the RGB image and depth at the input layer, forming four-channel 3D-aware data.
%
\cite{long2015fully} on the contrary, perform multi-modal fusion at the output layer, leading to a late fusion model.
To obtain more accurate and robust segmentation, a variety of models delicately design strategies for cross-level fusion, where they merge the complementary information in the middle level of the network \cite{park2017rdfnet, valada2019self, lin2017cascaded, feng2016local, liu2020learning, chen2018progressively, piao2019depth}.


\noindent\textbf{Confidence-aware Learning}
Confidence-aware learning (or uncertainty-aware learning) aims to estimate the uncertainty representing the quality of the data (aleatoric uncertainty) or the awareness of true model (epistemic uncertainty) \cite{what_uncertainty}. In this paper, as we mainly focus on modeling the quality of depth for camouflaged object detection, we only discuss the aleatoric uncertainty modeling strategies. \cite{kong2020sde} designs a network that yields a probabilistic distribution as output in order to capture such uncertainty. \cite{shen2021real} employs a teacher-student paradigm to distill the aleatoric uncertainty, where the teacher network generates multiple predicative samples by incorporating aleatoric uncertainty for the student network to learn. \cite{simple_scalable_uncertainty} uses an adversarial perturbation technique to generate additional training data for the model to capture the aleatoric uncertainty.

\noindent\textbf{Uniqueness of our solution} Different from the existing camouflaged object detection methods \cite{le2019anabranch, yan2020mirrornet, fan2020camouflaged, yunqiu_cod21} that rely only on the RGB images, we introduce the first depth-guided camouflaged object detection framework, with depth contribution estimation solutions to adaptively fuse model predictions from both the RGB branch and the RGB-D branch.
Although RGB-D data related segmentation models \cite{fan2020bbsnet,dmra_iccv2019} are widely studied, we claim that their multi-modal learning strategies based on \enquote{sensor depth} fails to explore the contribution of \enquote{generated depth}. On the contrary, they usually lead to worse performance compared with training only with RGB images (see Table \ref{tab:multi_modal_sensor_depth}). Our solution aims to produce a pixel-wise confidence map of each modal prediction, which is then treated as the weight to achieve multi-modal confidence-aware learning, leading to a better generated-depth contribution exploration.

\section{Our Method}
The original RGB image based COD training dataset is defined as $D=\{x_i,y_i\}_{i=1}^N$, where $x_i$ and $y_i$ are the input RGB image and the corresponding camouflage map, and $i$ indexes the images, $N$ is the size of $D$. In this paper, we study the contribution of depth for camouflaged object detection with a confidence-aware network shown in Fig.~\ref{fig:network_overview}.
To start our pipeline, we first generate the depth map $d$ with existing monocular depth estimation method \cite{MiDaS_Ranftl_2020_TPAMI}, leading to our RGB-D COD training dataset $D=\{x_i,d_i,y_i\}_{i=1}^N$. Then we introduce
an auxiliary depth estimation branch and a multi-modal confidence-aware learning loss function to effectively explore depth contribution for COD.

\subsection{Initial Depth Generation}
As there exists no RGB-D based camouflaged object detection dataset, we generate pseudo depth maps from existing monocular depth estimation methods.
We tried three
state-of-the-art monocular depth estimation methods (MiDaS\cite{MiDaS_Ranftl_2020_TPAMI}, Monodepth2\cite{monodepth2_Clement_2019_ICCV} and  FrozenPeople\cite{FrozenPeople_li_2019_CVPR}) to generate depth maps for our training
dataset.

\begin{figure}[thp]
   \begin{center}
   \begin{tabular}{{c@{ } c@{ } c@{ } c@{ }}}
   {\includegraphics[width=0.23\linewidth]{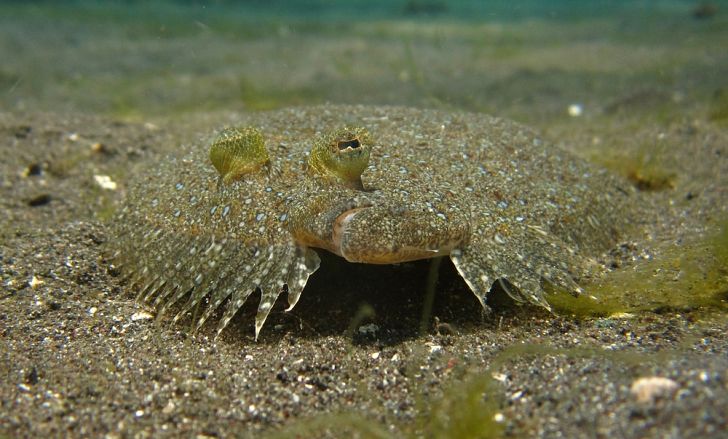}}&
    {\includegraphics[width=0.23\linewidth]{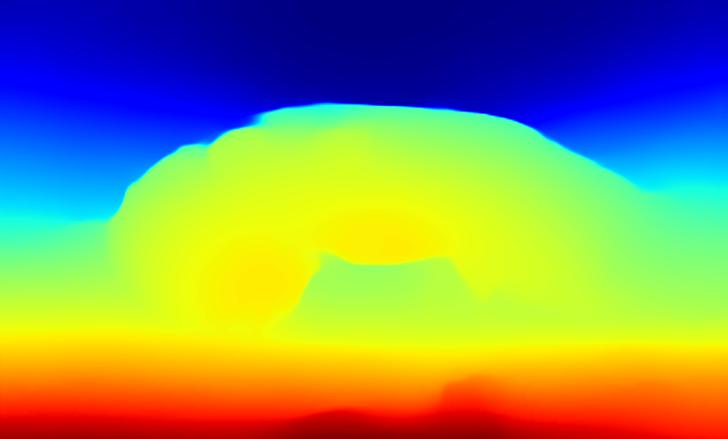}}&
    {\includegraphics[width=0.23\linewidth]{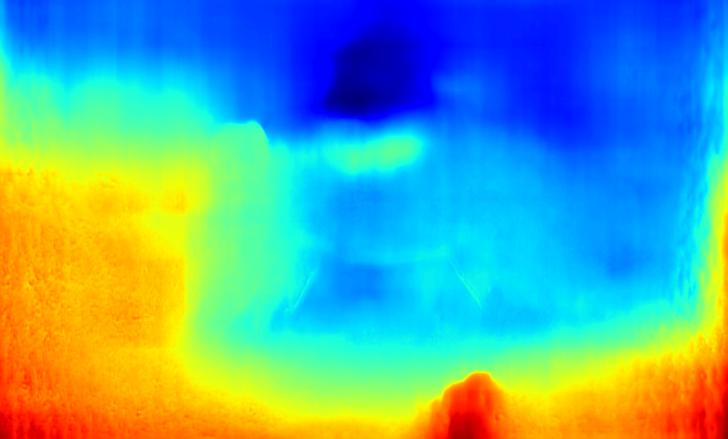}}&
    {\includegraphics[width=0.23\linewidth]{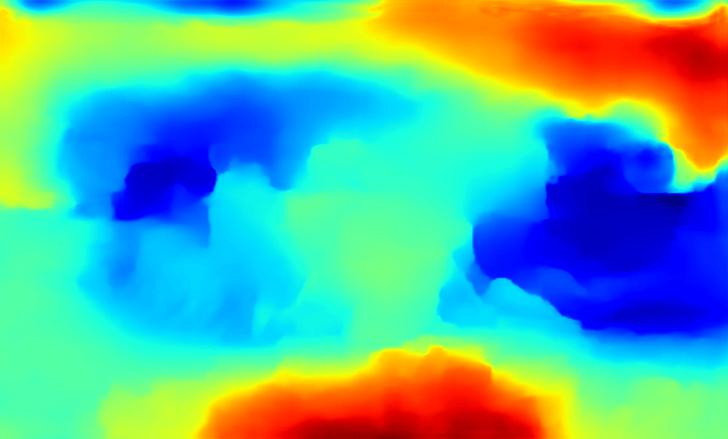}}\\
    {\includegraphics[width=0.23\linewidth]{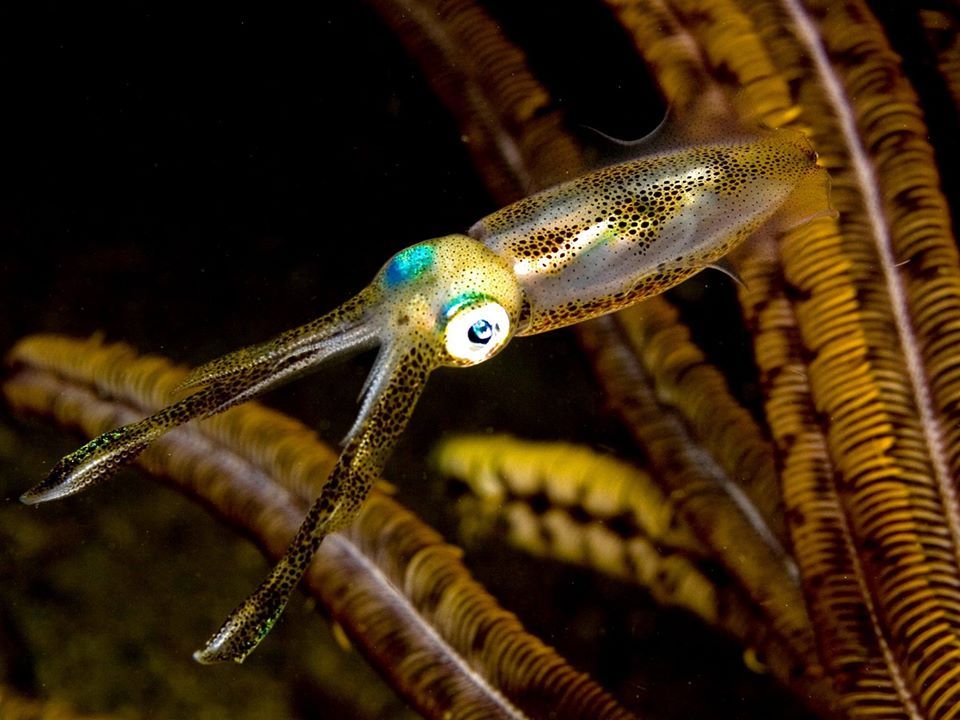}}&
    {\includegraphics[width=0.23\linewidth]{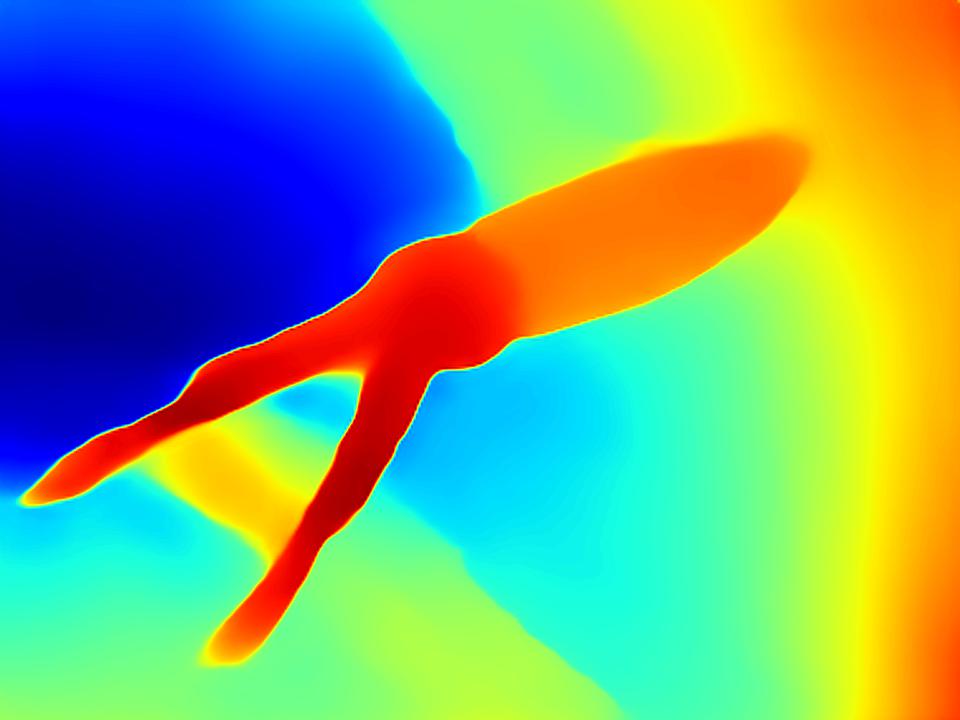}}&
    {\includegraphics[width=0.23\linewidth]{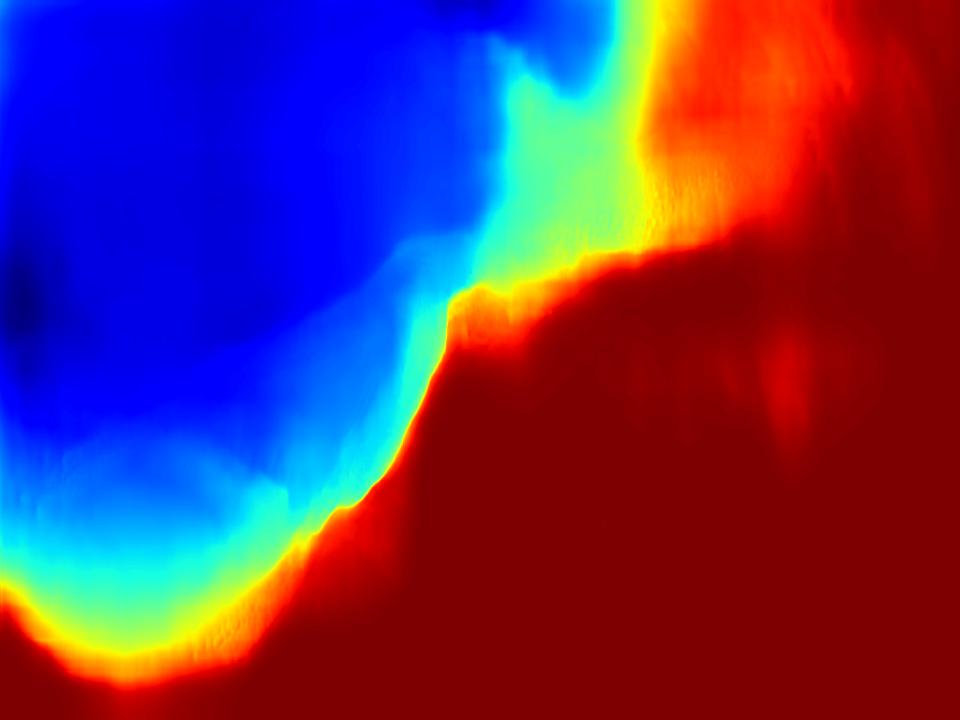}}&
    {\includegraphics[width=0.23\linewidth]{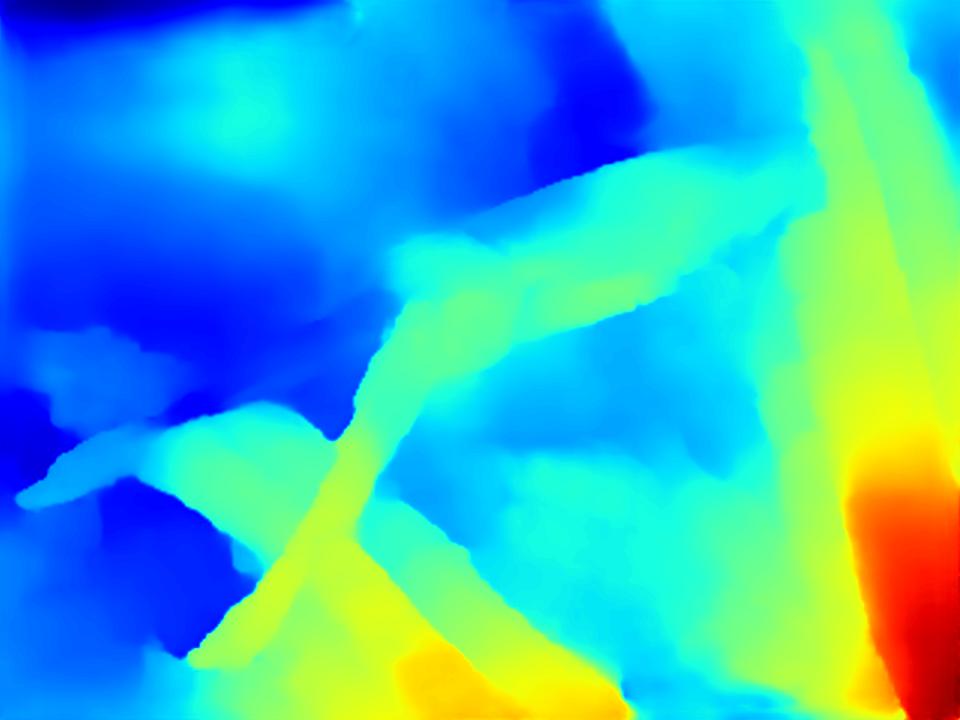}}\\
    \footnotesize{Image}  &\footnotesize{MiDaS~\cite{MiDaS_Ranftl_2020_TPAMI}} &\footnotesize{Monodepth2~\cite{monodepth2_Clement_2019_ICCV}}&
     \footnotesize{FrozenPeople~\cite{FrozenPeople_li_2019_CVPR}}\\
   \end{tabular}
   \end{center}
    \caption{Camouflage images and the generated depth maps. Three methods are used to generate depth maps. Overall, MiDaS~\cite{MiDaS_Ranftl_2020_TPAMI} has the best cross dataset performance and the strongest generalization ability. MonoDepth2~\cite{monodepth2_Clement_2019_ICCV} is trained only on KITTI dataset, which performs the worst in our scenario.} 
    \label{fig:depth_gen}
\end{figure}

Trained on 10 different RGB-D datasets, MiDaS~\cite{MiDaS_Ranftl_2020_TPAMI}
provides robust results across diverse scenes. 
Targeting autonomous driving, Monodepth2~\cite{monodepth2_Clement_2019_ICCV} has a good performance on the KITTI benchmark under the self-supervision, but its ability of domain adaption is to be investigated.
Since there are a large number of images with humans as camouflaged objects in our camouflage training dataset, we also tied FrozenPeople~\cite{FrozenPeople_li_2019_CVPR} with the single image configuration to generate depth for the characters in the scene.

Both MiDaS and Monodepth2 generate depth maps in the form of disparity, or inverse-depth, FrozenPeople generates depth values directly. We show the generated depth maps for our camouflaged object detection dataset in Fig.~\ref{fig:depth_gen}. Due to the visually better performance of MiDaS depth maps\footnote{The \enquote{depth map} represents \enquote{inverse-depth map} in the following.
}, we adopt MiDaS depth maps in our experiments.



\subsection{Depth Contribution Exploration}
\label{sec:depth_contribution_exploration}
Due to the domain gap, the generated depth map from monocular depth estimation methods may not be very accurate as shown in Fig.~\ref{fig:depth_gen}. Directly training with the less-accurate depth map cannot improve the model performance as the network will over-fit on the less-accurate depth map, leading to poor generalization ability (see Table \ref{tab:multi_modal_sensor_depth}).
Instead of directly using the depth as input, we design an auxiliary depth estimation branch, where the RGB-D feature fusion is achieved through multi-task learning (camouflaged object detection and monocular depth estimation) instead of multi-modal fusion within a single task.
Four main modules are included in our framework: 1) a RGB COD module to generate camouflage map from the RGB image; 2) a auxiliary depth estimation module to regress the depth map of image $x$ with $d$ as supervision; 3) a multi-modal fusion module to aggregate the intermediate features from the above two modules;
4) a multi-modal confidence-aware learning module via generative adversarial network (GAN) \cite{gan_raw} to adaptively weight the contribution of depth for model updating.

\noindent\textbf{RGB camouflaged object detection:}
We build the RGB camouflaged object detection network upon ResNet50~\cite{he2016deep} backbone, which maps the input image $x$ to
four stages features $f_{\theta_y}(x)=\{s_k\}_{k=1}^4$ of channel size 256, 512, 1024 and 2048 respectively, where $\theta_y$ is the parameter set of the backbone network. To obtain enlarged receptive field with less memory consumption, we feed each $s_k$ to a multi-scale dilated convolution block \cite{chen2017deeplab} to obtain new backbone features $f_{\theta_y}(x)=\{s'_k\}_{k=1}^4$ of channel size $C=32$. We further adopt decoder from \cite{MiDaS_Ranftl_2020_TPAMI} for high/low level feature aggregation. Specifically, we define prediction from the deterministic RGB branch as $f_{\beta_y}(f_{\theta_y}(x))$, where $\beta_y$ is the parameter set of the decoder (excluding the multi-scale dilated convolutional blocks). As we intend to model confidence of each modal (RGB and RGB-D), we change the RGB camouflaged object detection network to a probabilistic network via generative adversarial network (GAN) \cite{gan_raw}. In this way, we re-define model prediction of our stochastic prediction network as $f_{\beta_y}(f_{\theta_y}(x),z)$, where $z$ is the latent variable. To fuse $z$ with the backbone feature $f_{\theta_y}(x)$, we first tile $z$ to the same spatial size of $s'_4$, and then concatenate the tiled $z$ with $s'_4$, and feed it to another $3\times3$ convolutional layers of channel size $C=32$. We use this new feature instead of $s'_4$ in the decoder parametered with $\beta_y$.
Following the standard practice of GAN, $z$ is assumed to follow the standard normal distribution: $z\sim\mathcal{N}(0,\mathbf{I})$.

\noindent\textbf{Auxiliary depth estimation:} Given the generated depth map $d_i$ from \cite{MiDaS_Ranftl_2020_TPAMI}, we design auxiliary depth estimation branch to extract depth related features from the input image $x$. Specifically, we have a separate encoder with ResNet50 backbone for the auxiliary depth estimation branch with parameter set $\theta_d$. Similarly, multi-scale dilated connvolutional blocks are used for larger receptive filed to generate the depth backbone features $f_{\theta_d}(x)=\{s_k^{d'}\}_{k=1}^4$ of the same channel size $C=32$, and we adopt the same decoder structure in the RGB COD module to produce a one-channel depth map $d'=f_{\beta_d}(f_{\theta_d}(x))$.

\noindent\textbf{Multi-modal Fusion:}
Given intermediate backbone features $f_{\theta_y}(x,z)$ from the RGB COD branch and the auxiliary depth estimation branch $f_{\theta_d}(x)=\{s_k^{d'}\}_{k=1}^4$, the goal of the \enquote{Multi-modal Fusion} branch is to effectively fuse features for COD and features for MDE to
output a RGB-D camouflage map $f_{\alpha}(f_{\theta}(x),d,z^d)$, where $\theta=\{\theta_d,\theta_y\}$, $\alpha$ is the parameter set of the multi-modal fusion module, and $z^d$ is the latent variable for the RGB-D camouflage branch, and we also have $z^d\sim\mathcal{N}(0,\mathbf{I})$.
We concatenate each level of intermediate feature from RGB branch and depth branch, then we feed it to a $3\times3$ convolutions layer of channel size $C=32$, and defined the fused feature as $f_\theta^{rgbd}=f_\theta(x,d,z,z^d)$. We design a COD decoder within the multi-modal fusion module, which share the same structure as \cite{MiDaS_Ranftl_2020_TPAMI}. We define the RGB-D COD prediction as $f_{\alpha}$ for simplicity.


\noindent\textbf{Multi-modal Confidence-aware Learning:}
Recall that the RGB COD model produces RGB image based camouflage map $p(y|x,z)=\mathcal{N}(f_{\beta_y}(f_{\theta_y}(x),z),\epsilon_{rgb})$, where $\epsilon_{rgb}\sim\mathcal{N}(0,\sigma^2_{rgb})$ and $\sigma^2_{rgb}$ is the variance of the labeling noise, representing the inherent noise level from a generative model's perspective. Similarly, with the \enquote{Multi-modal Fusion} module, we obtain the RGB-D camouflage map $p(y|x,d,z^d)=\mathcal{N}(f_{\alpha},\epsilon_{rgbd})$, where $\epsilon_{rgbd}\sim\mathcal{N}(0,\sigma^2_{rgbd})$ also represents the inherent labeling noise.
Directly training the network with above two camouflage predictions and depth regression leads to inferior performance (see \enquote{A\_D} in Table \ref{tab:ablation_study}), as there exists no explicit constraint for depth quality evaluation. Although the proposed auxiliary depth estimation module achieves better depth exploration than other multi-modal learning strategies widely used in the existing RGB-D segmentation models \cite{fan2020bbsnet,dmra_iccv2019} (compare \enquote{A\_D} in Table \ref{tab:ablation_study} with multi-modal learning strategies in Table \ref{tab:multi_modal_sensor_depth}), there exists no explicit solution to weight the contribution of depth. Towards explicit depth contribution exploration, we first estimate confidence of each modal predictions, and use it as the weight within our multi-modal confidence-aware loss function.

\noindent\textbf{Prediction confidence estimation:}
Given stochastic prediction $p(y|x,z)$ from the RGB COD branch, and $p(y|x,d,z^d)$ from the RGB-D branch, we first perform multiple iterations of sampling in the latent space $z\sim\mathcal{N}(0,\mathbf{I})$ and $z^d\sim\mathcal{N}(0,\mathbf{I})$, and obtain a sequence of predictions $\{p(y|x,z_k)\}_{z_k}$ and $\{p(y|x,d,z^d_k)\}_{z^d_k}$, where $z_k$ and $z_k^d$ represent the random sample of the latent variable for the RGB COD branch and RGB-D COD branch respectively. For an ensemble based framework, the predictive uncertainty \cite{what_uncertainty} (the inverse confidence) captures the overall uncertainty of model prediction, which is defined as entropy of the mean prediction ($\mathbb{H}[.]$ in Fig.~\ref{fig:network_overview}). In this way, we obtain the confidence map of each modal prediction as:
\begin{equation}
\begin{aligned}
     \label{confidence_modals}
    &C_{\text{rgb}}=1-\mathbb{H}[\mathbb{E}_{z\sim\mathbb{N}(0,1)}p(y|x,z)],\\
    &C_{\text{rgbd}}=1-\mathbb{H}[\mathbb{E}_{z^d\sim\mathbb{N}(0,1)}p(y|x,d,z^d)].
\end{aligned}
\end{equation}
We show the inverse confidence maps (the uncertainty map) $U_{\text{rgb}}$ and $U_{\text{rgbd}}$ for RGB and RGB-D modal in Fig.~\ref{fig:our_solution_outputs}.

\noindent\textbf{Discriminator:}
As a generative adversarial network \cite{gan_raw},
we design an extra fully convolutional discriminator of the same structure as in \cite{hung2018adversarial}, and define it as $g_\gamma(\cdot)$, which is used to distinguish model predictions and the ground truth annotations. Specifically, the discriminator takes the concatenation of model prediction (or ground truth) and image as input, and identify it as \enquote{fake} (or real) with an all zero feature map $\mathbf{0}$ (or all one feature map $\mathbf{1}$).

\noindent\textbf{Train the model with multi-modal confidence-aware loss function:} Given the COD outputs
$p(y|x,z)$ and $p(y|x,d,z^d)$ from the RGB and RGB-D branch, and the corresponding confidence maps $C_{\text{rgb}}$ and $C_{\text{rgbd}}$, we introduce a multi-modal confidence-aware loss function to explicitly weight the depth contribution. The basic assumption is that if the model is confident about the prediction from one modal, then that modal should contribute more to the overall loss function, and vice versa. Inspired by \cite{georgecvpr2021}, our multi-modal confidence-aware loss function is defined:
\begin{equation}
    \label{multi_modal_confidence_loss}
    \mathcal{L}_{\text{cod}}=\omega_{\text{rgb}}\mathcal{L}_{\text{rgb}}+\omega_{\text{rgbd}}\mathcal{L}_{\text{rgbd}},
\end{equation}
where $\omega_{\text{rgb}}$ and $\omega_{\text{rgbd}}$ are the pixel-wise confidence-aware weights, defined as: $\omega_{\text{rgb}}=(C_{\text{rgb}})/(C_{\text{rgb}}+C_{\text{rgbd}})$ and $\omega_{\text{rgbd}}=(C_{\text{rgbd}})/(C_{\text{rgb}}+C_{\text{rgbd}})$ respectively. $\mathcal{L}_{\text{rgb}}=\mathcal{L}_{\text{rgb}}(f_{\beta_y}(f_{\theta_y}(x),z),y)$ and $\mathcal{L}_{\text{rgbd}}=\mathcal{L}_{\text{rgbd}}(f_{\alpha},y)$ are the task related loss functions, which are structure-aware loss functions \cite{wei2020f3net} in this paper.

In addition to Eq.~\ref{multi_modal_confidence_loss}, as a generative adversarial network with auxiliary depth estimation branch, we add the extra adversarial loss $\mathcal{L}_{\text{adv}}$ and depth regression loss $\mathcal{L}_{\text{depth}}$ to $\mathcal{L}_{\text{cod}}$, and define it as our generator loss $\mathcal{L}_{\text{gen}}$:
\begin{equation}
    \label{generator_loss}
    \mathcal{L}_{\text{gen}}=\mathcal{L}_{\text{cod}}+\mathcal{L}_{\text{depth}}+\lambda\mathcal{L}_{\text{adv}},
\end{equation}
where $\lambda$ is the hyper-parameter to weight the contribution of the adversarial loss, and empirically we set $\lambda=0.1$. The adversarial loss $\mathcal{L}_{\text{adv}}$ is defined as:
\begin{equation}
\begin{aligned}
    \label{adversarial_loss}
    \mathcal{L}_{\text{adv}}&=\mathcal{L}_{ce}(g_\gamma(x,f_{\beta_y}(f_{\theta_y}(x),z)),\mathbf{1})\\
    &+\mathcal{L}_{ce}(g_\gamma(x,f_{\beta}(f_{\theta}(x),d,z)),\mathbf{1}),
\end{aligned}
\end{equation}
with $\mathcal{L}_{ce}$ as the binary cross-entropy loss. The extra adversarial loss aims to fool the discriminator to predict \enquote{real} with generator camouflage prediction as input. The depth regression loss $\mathcal{L}_{\text{depth}}$ is defined as the weighted sum of the point-wise $L_1$ loss and the SSIM loss \cite{ssim_depth}:
\begin{equation}
    \label{depth_loss}
    \mathcal{L}_{\text{depth}} = (1-\lambda) \frac{1}{n}\sum_{i=1}^n \left| d-d' \right| + \lambda \frac{1-SSIM(d, d')}{2},
\end{equation}
and empirically we set $\lambda=0.85$ \cite{ssim_depth}.

Then the discriminator is updated via:
\begin{equation}
\begin{aligned}
\label{discriminator_loss}
    \mathcal{L}_{\text{dis}}&=\mathcal{L}_{ce}(\mathcal{L}_{ce}(g_\gamma(x,y),\mathbf{1})\\
    &+ \mathcal{L}_{ce}(\mathcal{L}_{ce}(g_\gamma(x,f_{\beta_y}(f_{\theta_y}(x),z)),\mathbf{0})\\
    &+\mathcal{L}_{ce}(g_\gamma(x,f_{\alpha}),\mathbf{0})),\\
\end{aligned}
\end{equation}
where the discriminator aims to correctly distinguish model predictions from the ground truth, and this is what \cite{gan_raw} defined as the \enquote{minimax game}. We show the complete learning pipeline of our method in Algorithm \ref{alg_depth_guided_rgbd_cod}.

\begin{algorithm}[H]
\small
\caption{Depth Contribution Exploration for COD}
\textbf{Input}: \\
(1) Training dataset $D=\{x_i,d_i,y_i\}_{i=1}^N$, where the depth map $d_i$ is pre-computed from \cite{MiDaS_Ranftl_2020_TPAMI}.
\\
(2) The maximal number of learning epochs $E$. \\
\textbf{Output}: 
Model parameter $\theta=\{\theta_d,\theta_y\}$, $\beta=\{\beta_d,\beta_y\}$ and $\alpha$ for the camouflage generator, and $\gamma$ for the discriminator;
\begin{algorithmic}[1]
\For{$t \leftarrow  1$ to $E$}
\State Generate camouflage map $f_{\beta_y}(f_{\theta_y}(x),z)$ and $f_{\alpha}$ from the RGB and RGB-D COD branches.
\State Obtain monocular depth $f_{\beta_d}(f_{\theta_d}(x_i))$ of the input $x_i$.
\State Perform $T=5$ iterations of sampling from the latent space: $z\sim\mathcal{N}(0,\mathbf{I})$, $z^d\sim\mathcal{N}(0,\mathbf{I})$, and obtain $\{p(y|x,z_k)\}_{z_k}$ and $\{p(y|x,d,z_k)\}_{z^d_k}$, where $z_k$ and $z^d_k$ represent the random samples of the latent variables.
\State Compute confidence of each model $C_{\text{rgb}}$ and $C_{\text{rgbd}}$.
\State Obtain the pixel-wise confidence weights $\omega_{\text{rgb}}=(C_{\text{rgb}})/(C_{\text{rgb}}+C_{\text{rgbd}})$ and $\omega_{\text{rgbd}}=(C_{\text{rgbd}})/(C_{\text{rgb}}+C_{\text{rgbd}})$;
\State Define the multi-modal confidence-aware loss function $\mathcal{L}_{\text{cod}}=\omega_{\text{rgb}}\mathcal{L}_{\text{rgb}}+\omega_{\text{rgbd}}\mathcal{L}_{\text{rgbd}}$, and obtain generator loss $\mathcal{L}_{\text{gen}}$.
\State Update $\theta=\{\theta_d,\theta_y\}$, $\beta=\{\beta_d,\beta_y\}$ and $\alpha$ via $\mathcal{L}_{\text{gen}}$.
\State Compute discriminator loss $\mathcal{L}_{\text{dis}}$, and update $\gamma$ with $\mathcal{L}_{\text{dis}}$.
\EndFor
\end{algorithmic} 
\label{alg_depth_guided_rgbd_cod}
\end{algorithm}

\begin{table*}[t!]
  \centering
  \footnotesize
  \renewcommand{\arraystretch}{1.1}
  \renewcommand{\tabcolsep}{0.9mm}
  \caption{Performance comparison with benchmark COD models.
  \enquote{BkB}: the backbone model, and \enquote{R50} is the ResNet50 backbone \cite{resnet_he}, \enquote{R2\_50} is the Res2Net50 backbone \cite{res2net}. \enquote{Size}: size of the training and testing images.}
  \begin{tabular}{lccc|cccc|cccc|cccc|cccc}
  \hline
  &&&&\multicolumn{4}{c|}{CAMO~\cite{le2019anabranch}}&\multicolumn{4}{c|}{CHAMELEON~\cite{Chameleon2018}}&\multicolumn{4}{c|}{COD10K~\cite{fan2020camouflaged}}&\multicolumn{4}{c}{NC4K~\cite{yunqiu_cod21}} \\
    Method &BkB&Year&Size& $S_{\alpha}\uparrow$&$F_{\beta}\uparrow$&$E_{\xi}\uparrow$&$\mathcal{M}\downarrow$& $S_{\alpha}\uparrow$&$F_{\beta}\uparrow$&$E_{\xi}\uparrow$&$\mathcal{M}\downarrow$ &  $S_{\alpha}\uparrow$ & $F_{\beta}\uparrow$ & $E_{\xi}\uparrow$ & $\mathcal{M}\downarrow$ & $S_{\alpha}\uparrow$
    & $F_{\beta}\uparrow$ & $E_{\xi}\uparrow$ & $\mathcal{M}\downarrow$  \\
  \hline
   &&&\multicolumn{16}{c}{RGB COD Models} \\ \hline
  SINet~\cite{fan2020camouflaged}
  & R50 & 2020 &352& .745 & .702 & .804 & .092 & .872 & .827 & .936 & .034 & .776 & .679 & .864 & .043 & .810 & .772 & .873 & .057 \\
    SINet-V2~\cite{fan2021concealed}
  & R2\_50 & 2021 &352& .820 & .782 & .882 & .070 & .888 & .835 & .942 & .030 & .815 & .718 & .887 & .037 & .847 & .805& .903 & .048\\ 
  LSR~\cite{yunqiu_cod21}
  & R50 & 2021 &352 & .793 & .725 & .826 & .085 & .893 & .839 & .938 & .033 & .793 & .685 & .868 & .041 & .839 & .779 & .883 & .053  \\
  MGL~\cite{zhai2021Mutual}
  & R50 & 2021 &473& .775 & .726 & .812 & .088 & .893 & .834 & .918 & .030 & .814 & .711 & .852 & .035 & .833 & .782 & .867 & .052\\
  PFNet~\cite{mei2021Ming}
  & R50 & 2021 &416 & .782 & .744 & .840 & .085 & .882 & .826 & .922 & .033 & .800 & .700 & .875 & .040 & .829 & .782 & .886 & .053 \\
  UJTR~\cite{Yang_2021_ICCV_COD_Uncertainty}
  & R50 & 2021 &473& .785 & .686 & .859 & .086 & .888 & .796 & .918 & .031 & .818 & .667 & .850 & .035 & .839 & .786 & .873 & .052 \\
  \hline
   &&&\multicolumn{16}{c}{RGB-D COD Models} \\ \hline
   UCNet~\cite{ucnet_sal} &R50&2020&352& .729 & .672 & .785 & .101 & .869 & .823 & .924 & .039 & .738 & .611 & .825 & .052 & .784 & .728 & .849 & .066  \\ 
  BBSNet~\cite{fan2020bbsnet} &R50&2020&352& .776 & .689 & .786 & .093 & .864 & .768 & .872 & .049 & .782 & .633 & .836 & .050  & .825 & .745 & .859 & .062    \\ 
  JL-DCF~\cite{fu2020jl} &R50&2020&352& .772 & .670 & .777 & .102 & .857 & .734 & .867 & .052 & .749 & .581 & .789 & .053 & .788 & .713 & .819 & .072   \\ 
 SSF~\cite{ssf_cvpr2020} &VGG16&2020&352& .748 & .643 & .782 & .116 & .866 & .788 & .904 & .045 & .729 & .593 & .778 & .055 & .770 & .689 & .827 & .069  \\
  \hline
   Ours  &R50&2021&352& .794 &.759 & .853 & .077 & .885 &.836 & .941 & .032 & .801 &.705 & .882 & .037 & .837 &.798 & .899 & .049  \\
    Ours &R50&2021&416& .798 &.762 & .860 & .077 & .889 &.842 & .947 & .029 & .807 &.712 & .881 & .037 & .842 &.802 & .901 & .047  \\ 
    Ours &R50&2021&480& .803 &.770 & .858 & .078 & .891 &.846 & .939 & .028 & .820 &.737 & .894 & .033 & .845 &.809 & .903 & .046  \\ 
    Ours&R2\_50&2021&352& .819 &.798 & .881 & .069 & .895 &.856 & .951 & .027 & .829 &.751 & .903 & .032 & .855 &.825 & .910 & .042  \\
   \hline
  \end{tabular}
  \label{tab:benchmark_model_comparison_rgb}
\end{table*}

\begin{figure*}[thp]
   \begin{center}
   \begin{tabular}{{c@{ } c@{ } c@{ } c@{ } c@{ } c@{ } c@{ } c@{ }}}
   {\includegraphics[width=0.115\linewidth]{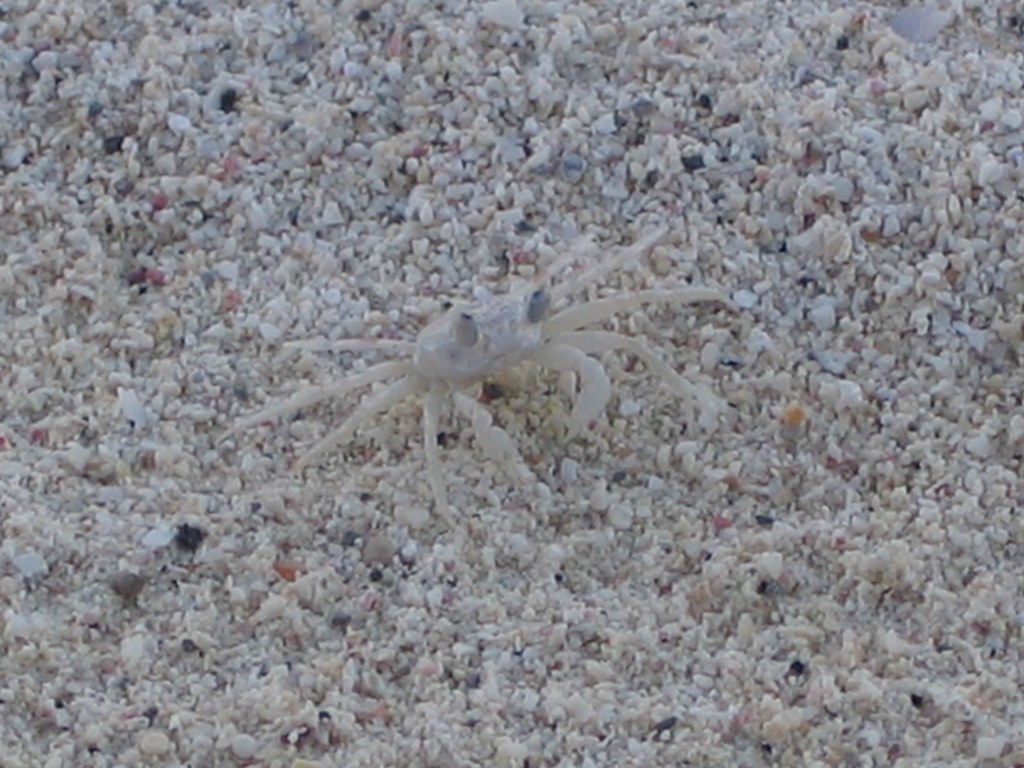}}&
    {\includegraphics[width=0.115\linewidth]{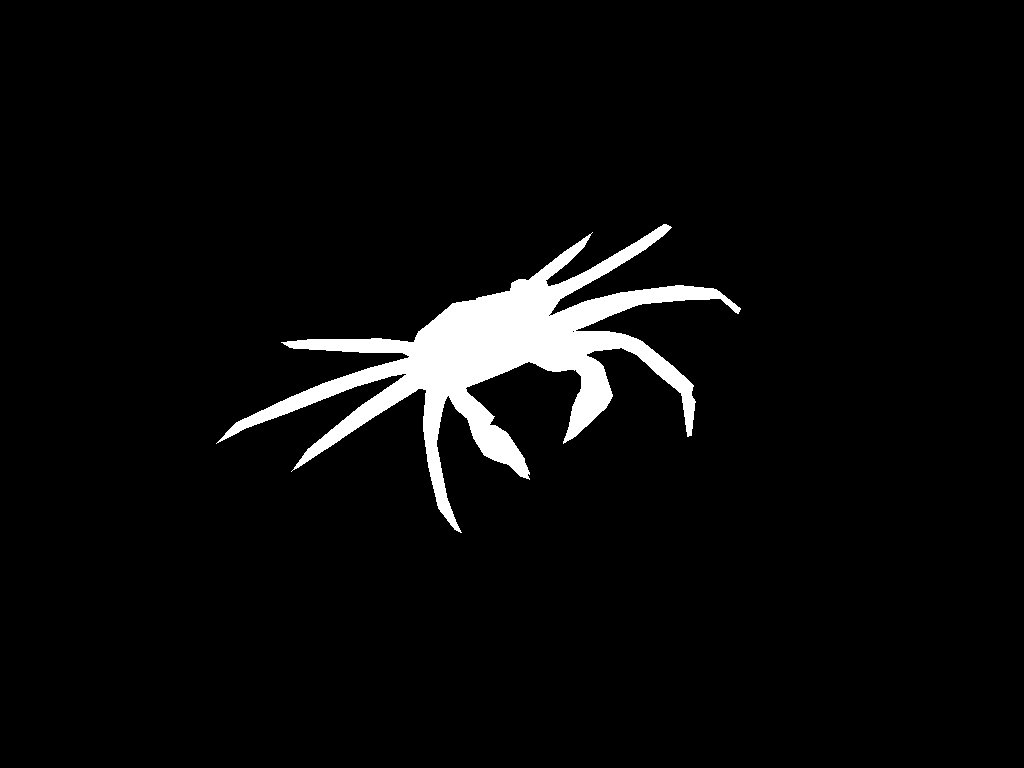}}&
    {\includegraphics[width=0.115\linewidth]{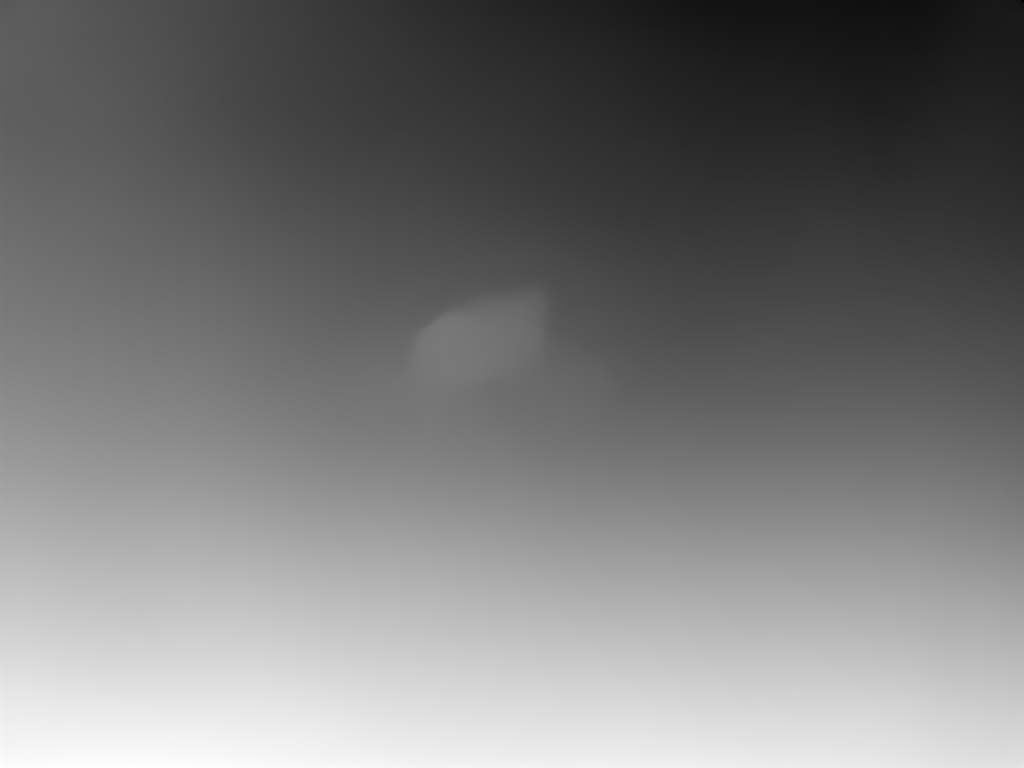}}&
    {\includegraphics[width=0.115\linewidth]{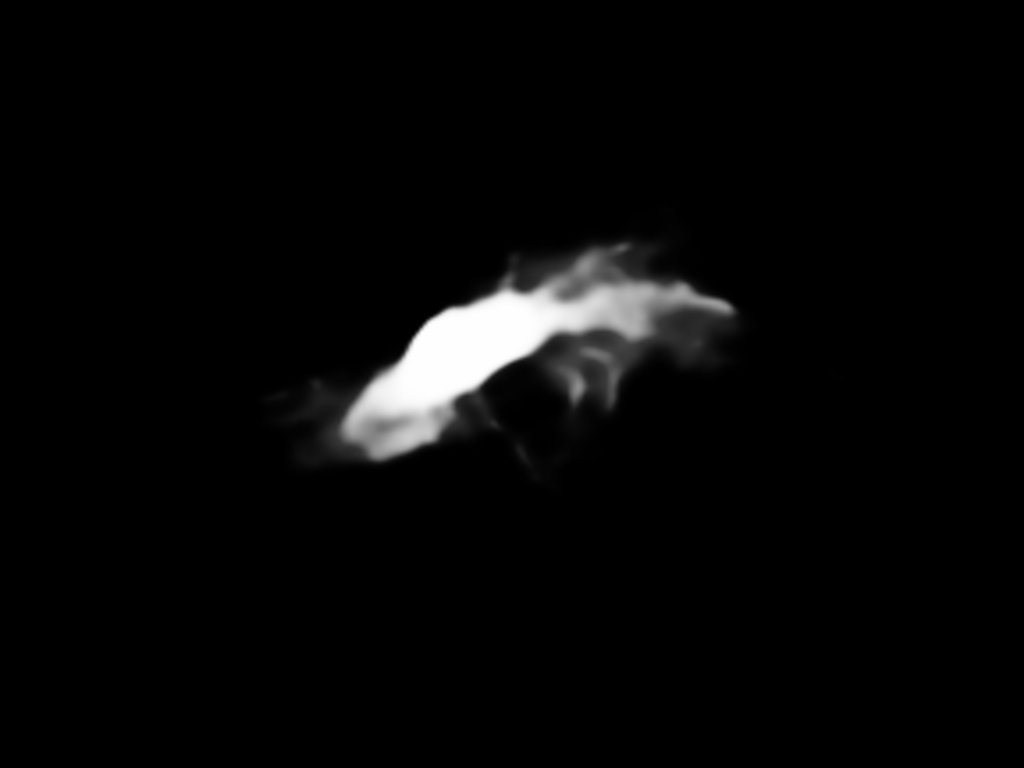}}&
    {\includegraphics[width=0.115\linewidth]{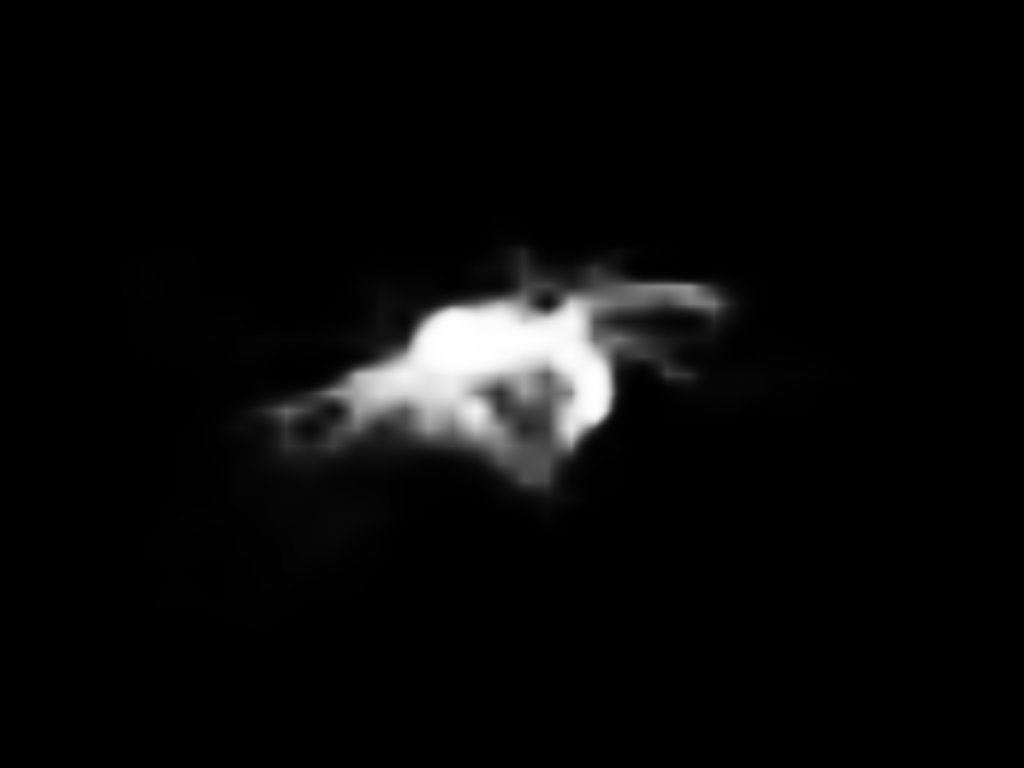}} &
    {\includegraphics[width=0.115\linewidth]{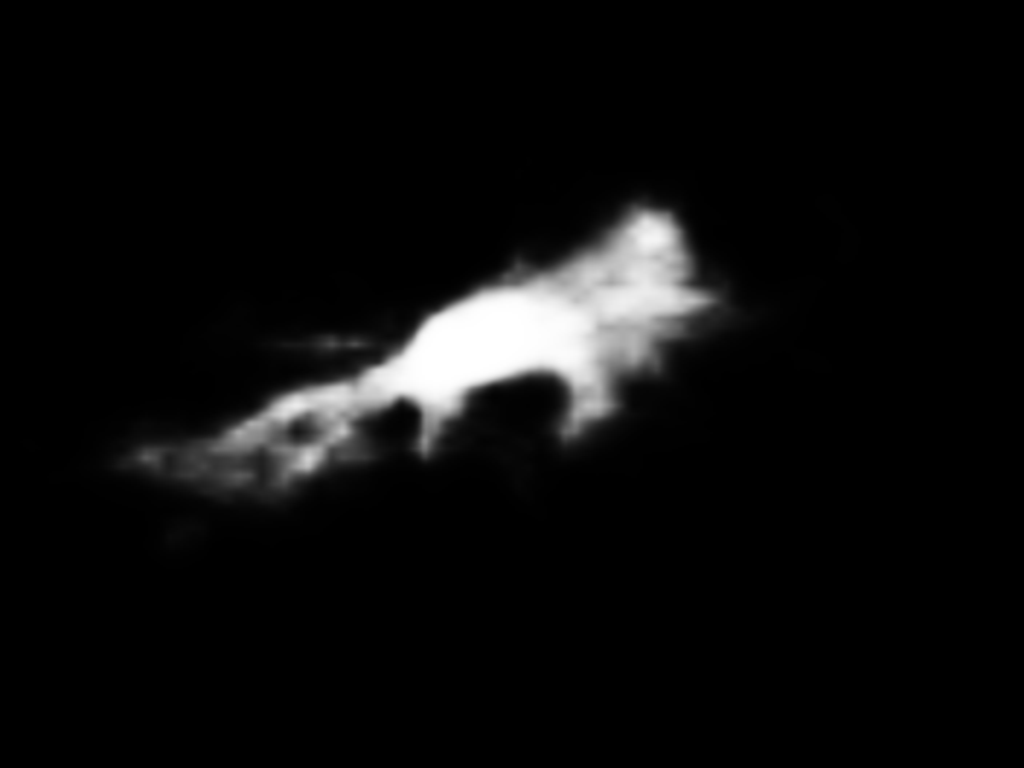}}&
    {\includegraphics[width=0.115\linewidth]{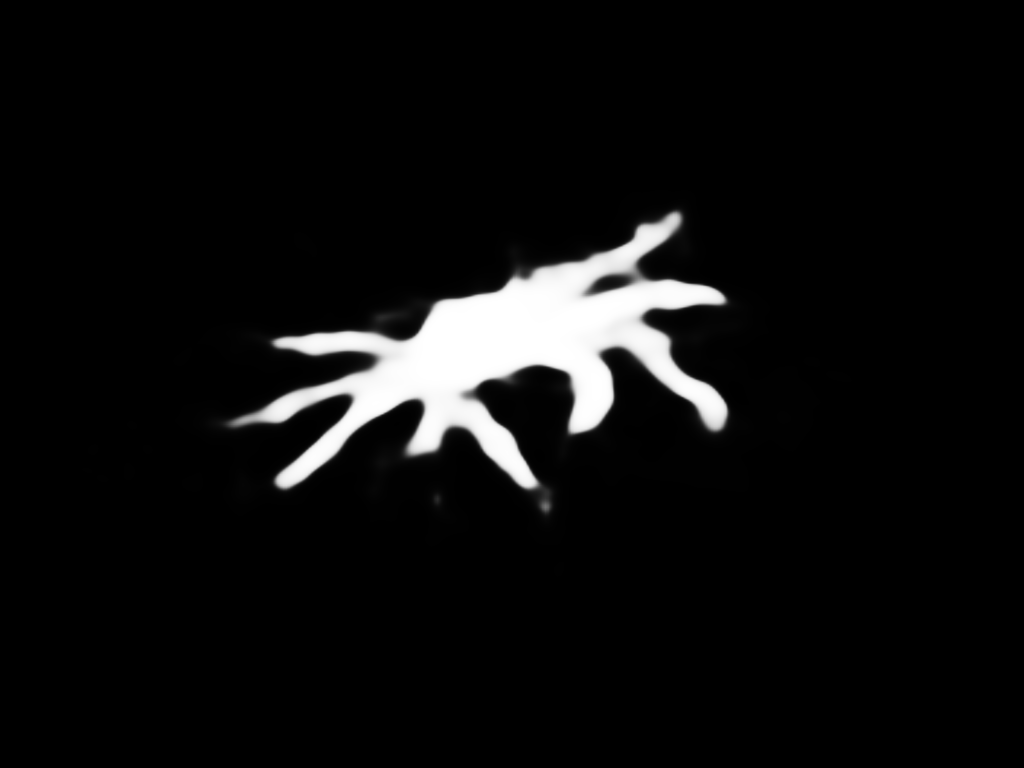}}&
    {\includegraphics[width=0.115\linewidth]{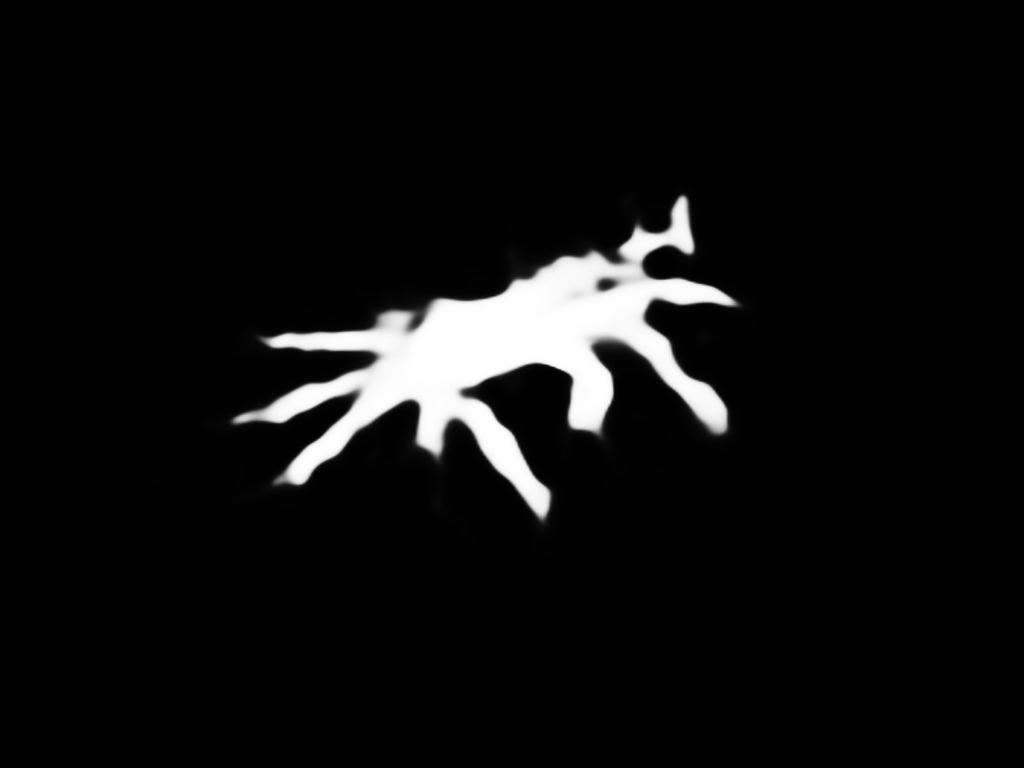}}\\
     {\includegraphics[width=0.115\linewidth]{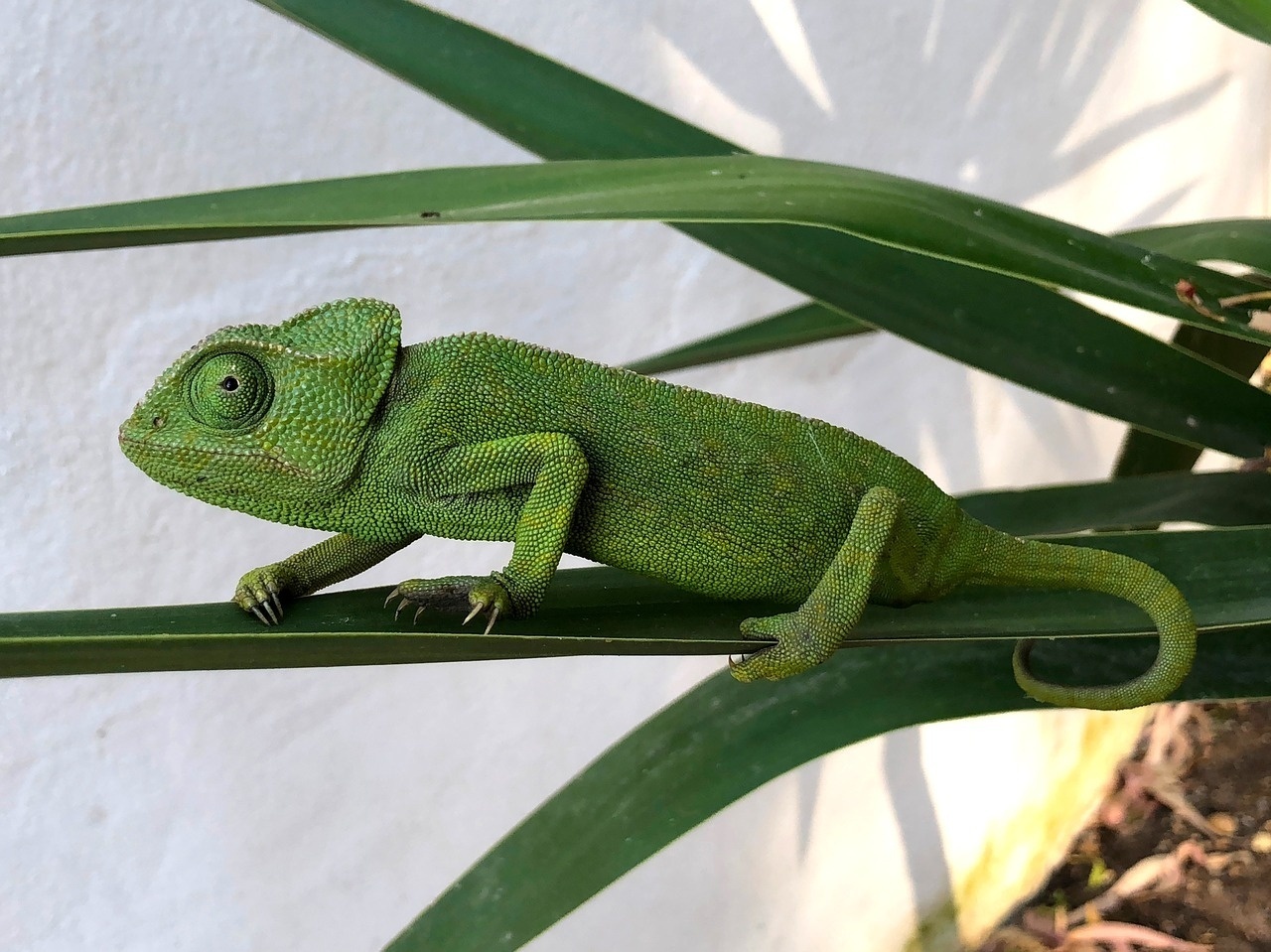}}&
    {\includegraphics[width=0.115\linewidth]{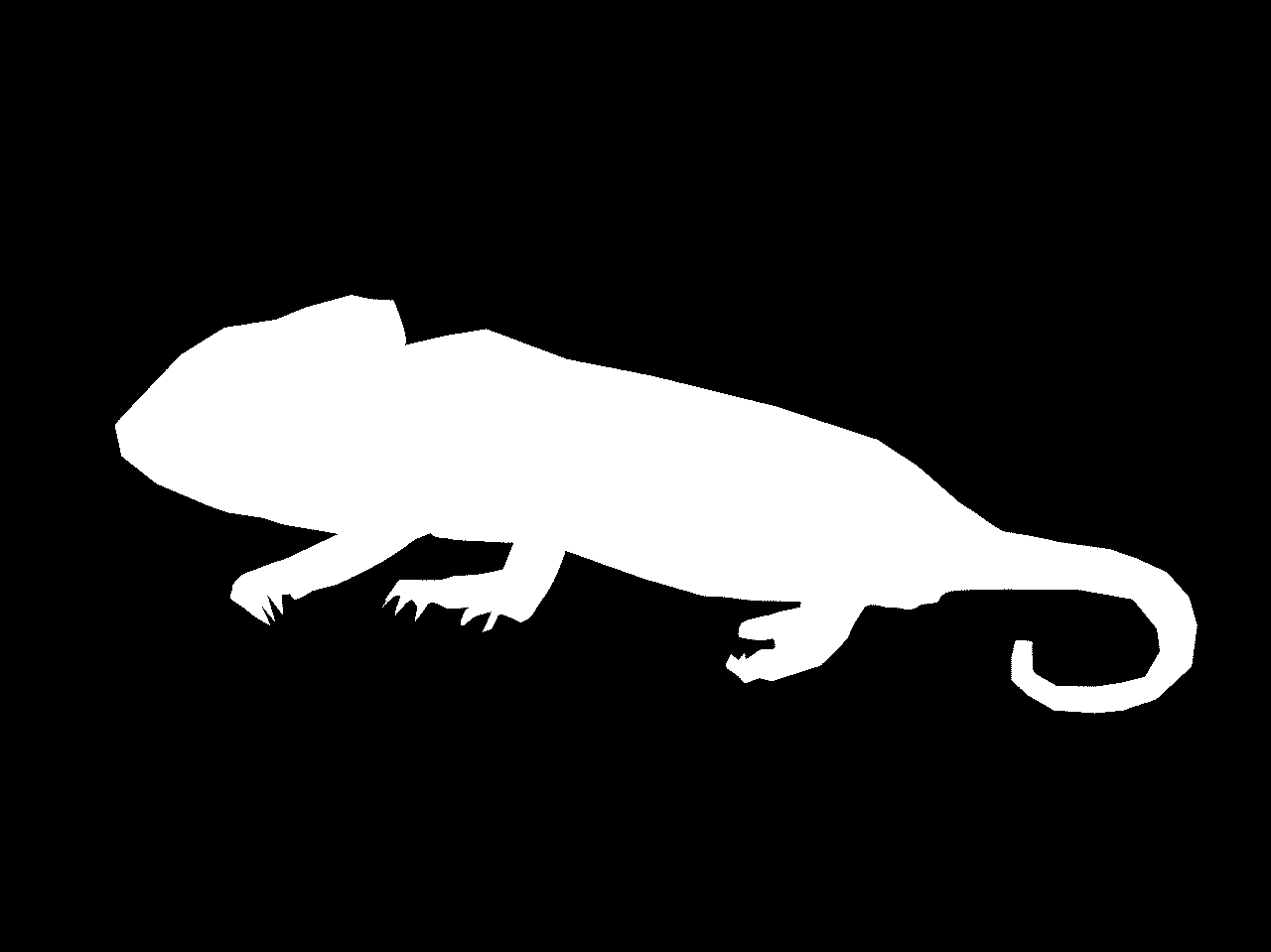}}&
    {\includegraphics[width=0.115\linewidth]{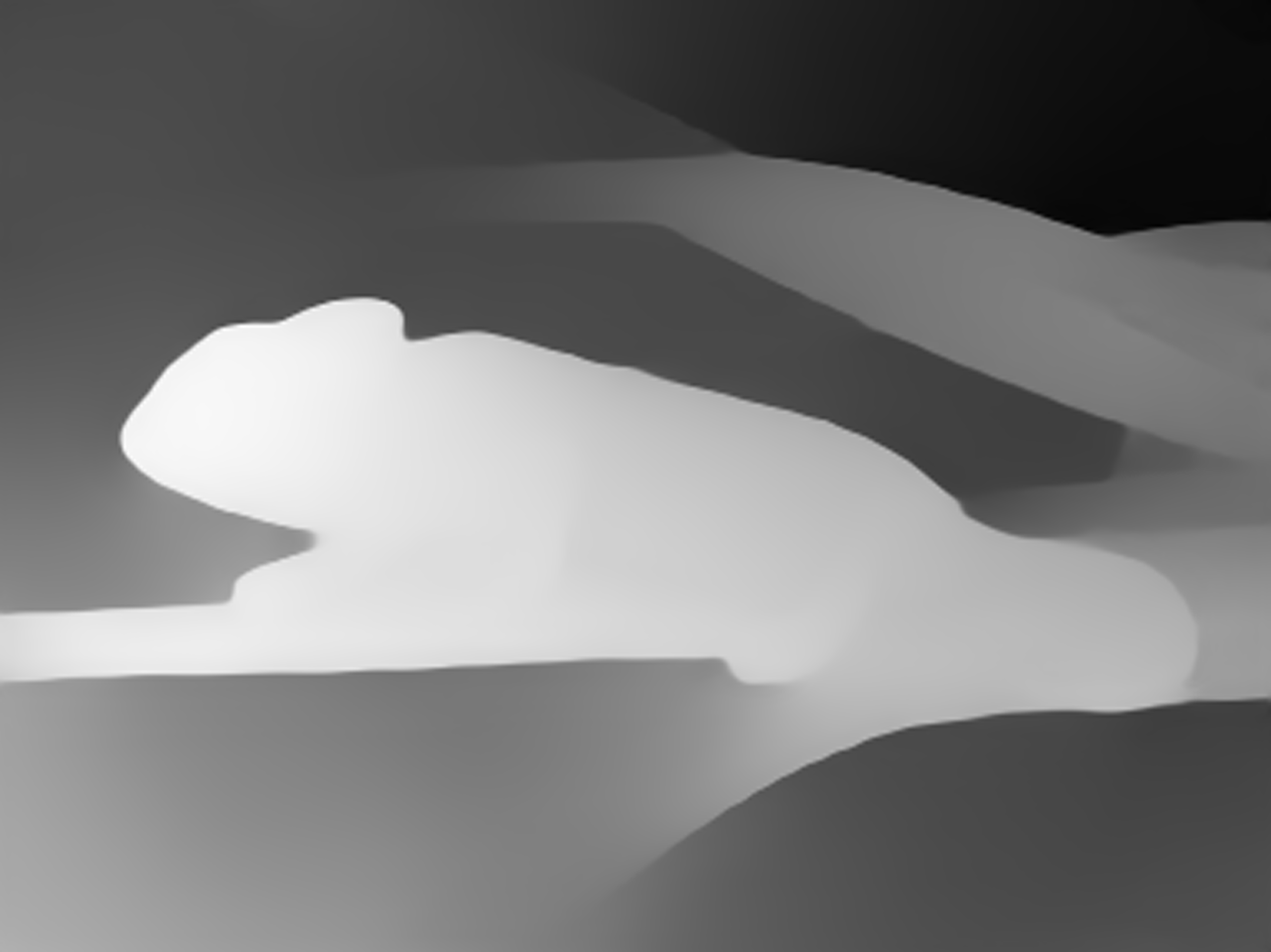}}&
    {\includegraphics[width=0.115\linewidth]{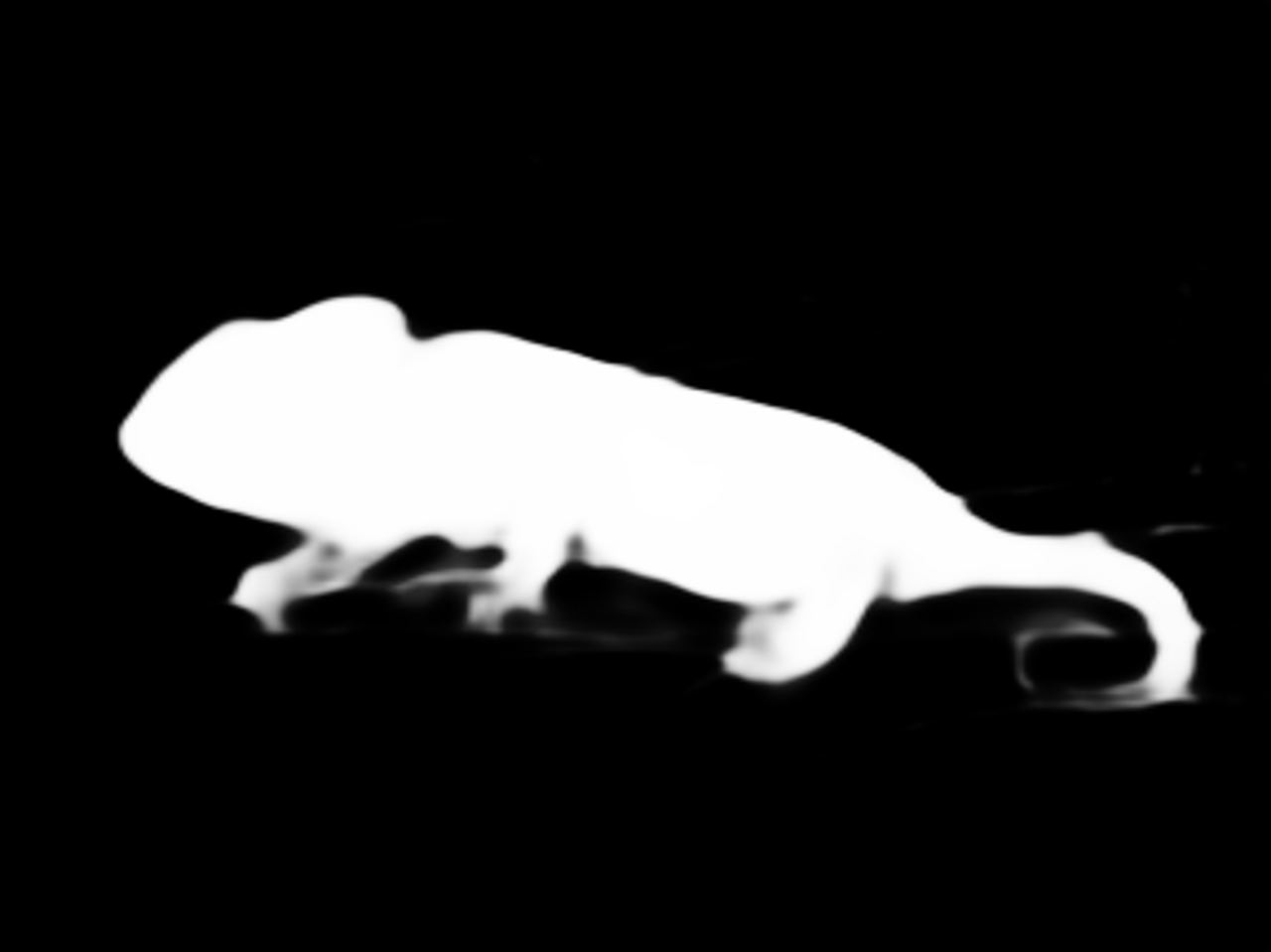}}&
    {\includegraphics[width=0.115\linewidth]{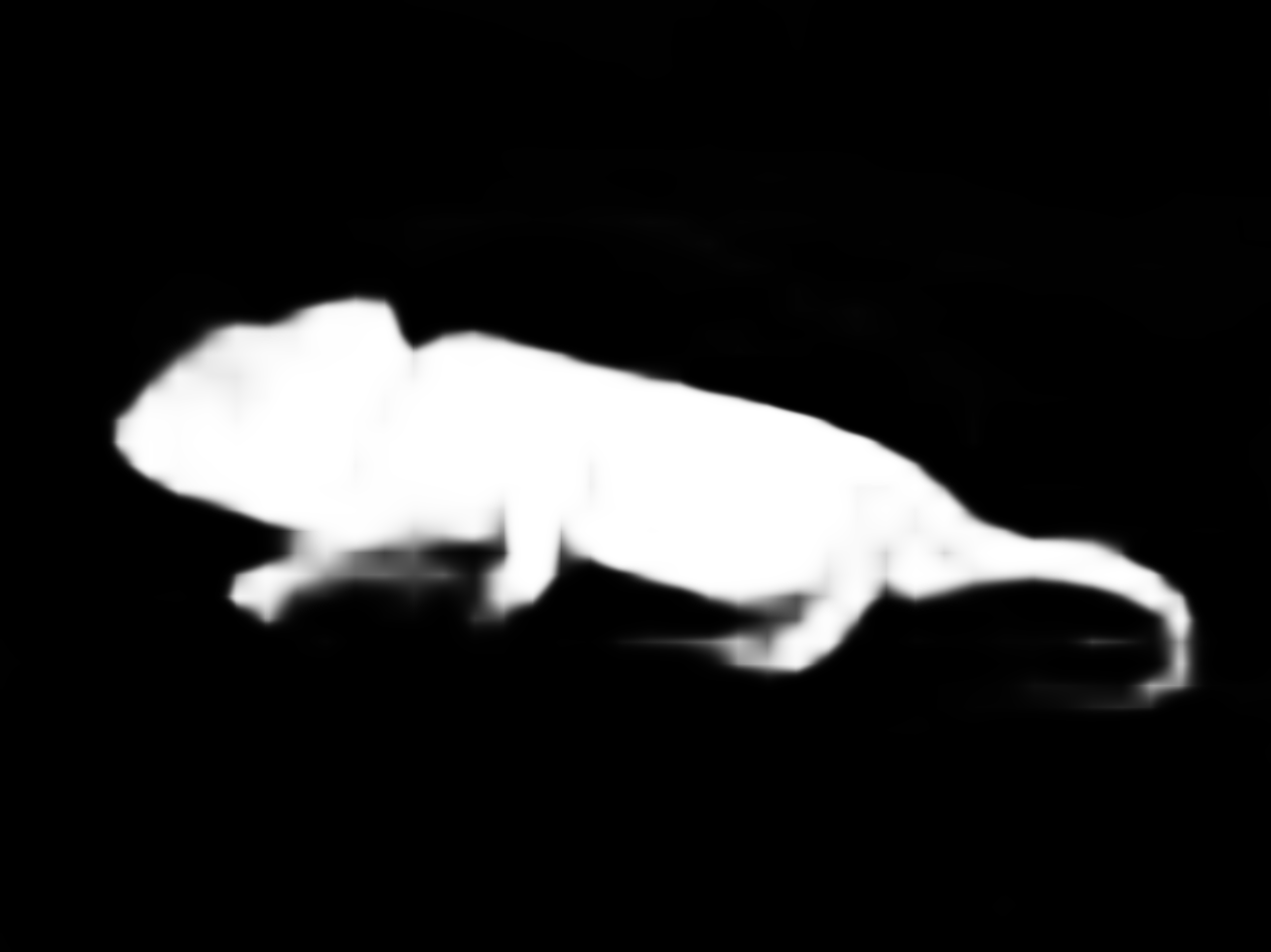}} &
    {\includegraphics[width=0.115\linewidth]{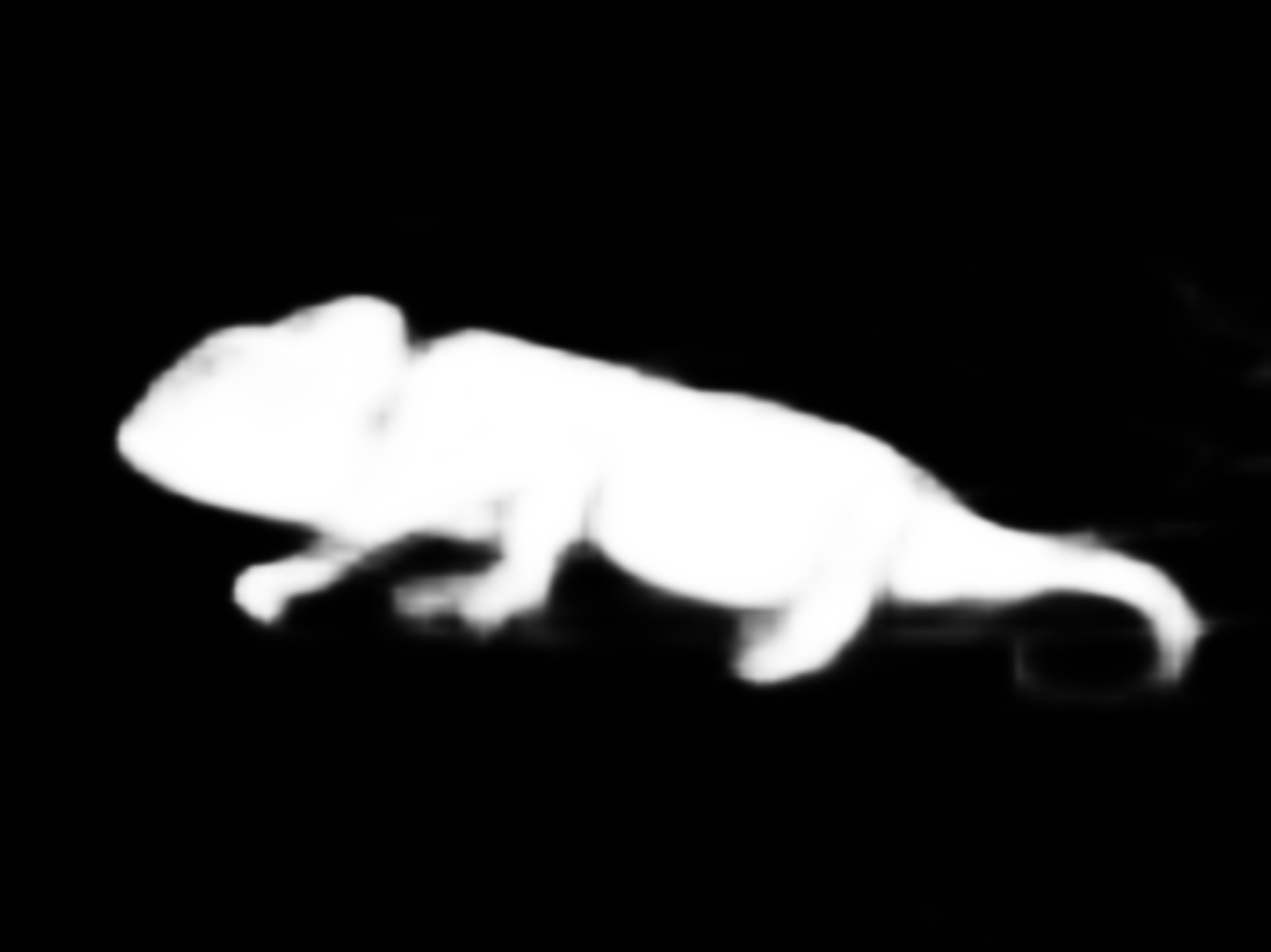}}&
    {\includegraphics[width=0.115\linewidth]{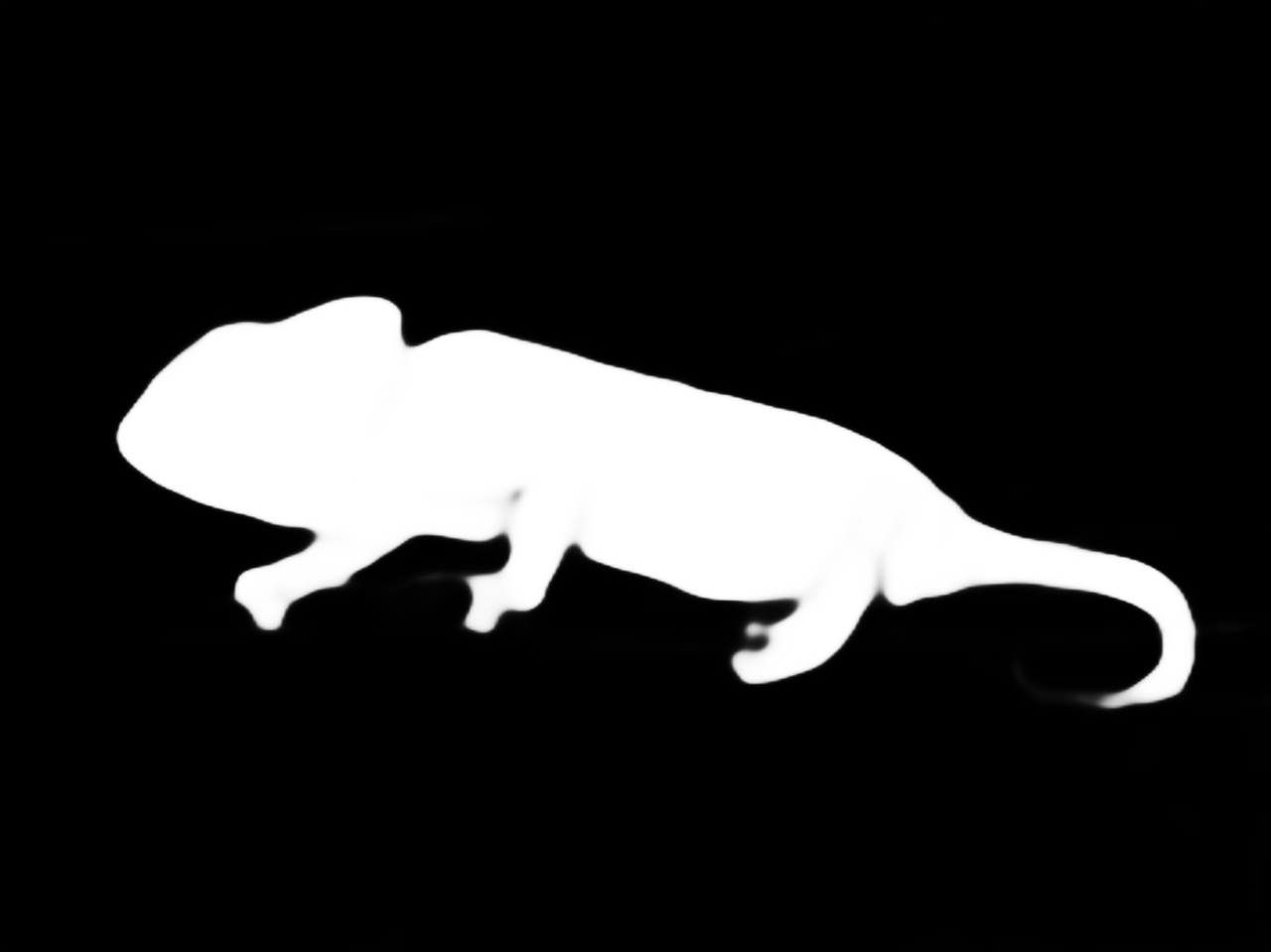}}&
    {\includegraphics[width=0.115\linewidth]{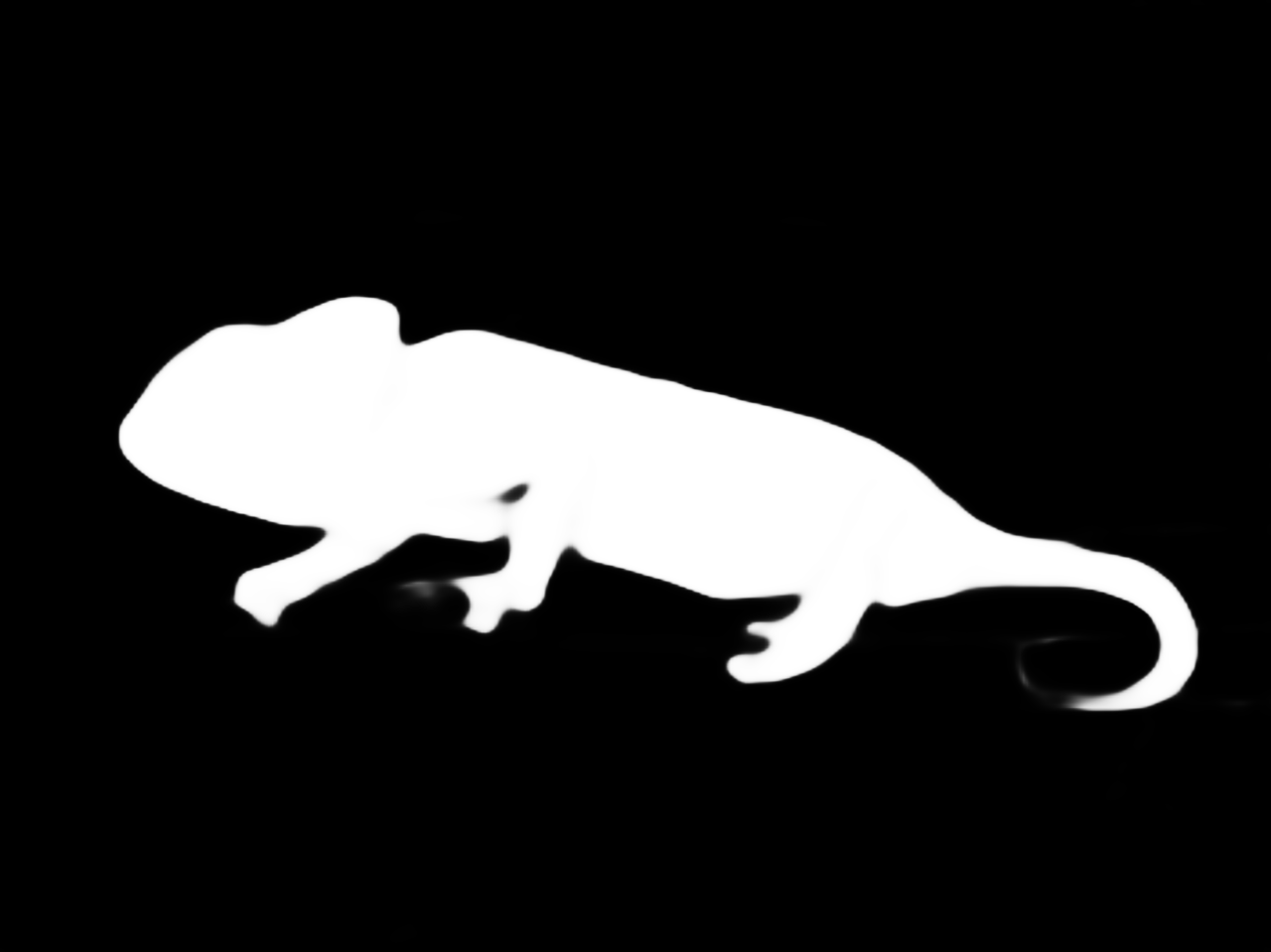}}\\
    \footnotesize{Image} 
    &\footnotesize{GT} 
    &\footnotesize{Depth}
    &\footnotesize{SINet~\cite{fan2020camouflaged}} &\footnotesize{MGL~\cite{zhai2021Mutual}} &\footnotesize{UJTR~\cite{Yang_2021_ICCV_COD_Uncertainty}} 
    &\footnotesize{Ours\_{RGB}} &\footnotesize{Ours\_{RGBD}}\\
   \end{tabular}
   \end{center}
    \caption{Visual comparison of our predictions with the benchmark techniques.} 
    \label{fig:visual_comparison}
\end{figure*}

\section{Experimental Results}
\subsection{Setup}
\noindent\textbf{Datasets}
We train our method with benchmark COD training dataset, which is the combination of 3,040 images from COD10K training dataset \cite{fan2020camouflaged} and 1,000 images from CAMO training dataset \cite{le2019anabranch}. We then test model performance on four benchmark camouflage testing datasets, including CAMO \cite{le2019anabranch} of size 250, CHAMELEON of size 76, COD10K testing set of size 2,026, and the newly released NC4K dataset \cite{yunqiu_cod21} of size 4,121.

\noindent\textbf{Evaluation Metrics}
We adopt four widely used metrics for performance evaluation, including 1)~Mean Absolute Error ($\mathcal{M}$); 2)~Mean F-measure ($F_\beta$); 3)~Mean E-measure ($E_\xi$) \cite{fan2018}; 4)~S-measure ($S_{\alpha}$) \cite{fan2017structure}. Details of those evaluation metrics are introduced in the supplementary materials.


\noindent\textbf{Implementation Details}
We train our model with Pytorch, where ResNet-50 is chosen as backbone, which is initialized with 
weights trained on ImageNet, and other newly added layers are initialized by default. We resize all the images and ground truth to $352\times352$ for both training and testing. The maximum epoch is 50. The initial learning rates is $2.5 \times 10^{-5}$. The whole training takes 17 hours with batch size 6 on one NVIDIA GTX 2080Ti GPUs.

\subsection{Performance comparison}
\noindent\textbf{Quantitative comparison:} We compare our results with the benchmark models in Table \ref{tab:benchmark_model_comparison_rgb}. As both training image size and backbone networks are important factors for model performance, for fair comparison, we train with different backbones and different training image sizes following the settings of existing solutions. Due to extra depth information, we also generate benchmark RGB-D COD models by training RGB-D saliency detection models with our RGB-D COD dataset.
Specifically, we re-train UCNet \cite{ucnet_sal}, BBSNet \cite{fan2020bbsnet}, JL-DCF \cite{fu2020jl} and SSF \cite{ssf_cvpr2020}) with our RGB-D camouflaged object detection dataset.
The consistent better performance the RGB COD benchmark models compared with the re-trained RGB-D COD models explains that the existing sensor depth based RGB-D fusion strategies work poorly within our generated depth scenario. The main reason lies in the low quality depth map due to domain gap of monocular depth estimation dataset and our camouflaged object detection dataset. Table \ref{tab:benchmark_model_comparison_rgb} shows that, with the proposed depth contribution exploration techniques, we obtain consistently improved performance compared with both the RGB COD models and the re-trained RGB-D COD models.


\noindent\textbf{Qualitative comparison:} In Fig.~\ref{fig:visual_comparison}, we show predictions of benchmark techniques and our method, where we show prediction from both the RGB COD branch (\enquote{Ours\_{RGB}}) and the RGB-D COD branch (\enquote{Ours\_{RGBD}}). The better camouflage maps from both branches
further explain effectiveness of our solutions.

\noindent\textbf{Running time comparison:} The parameter number of our camouflage generator
is 101M, which is larger than some RGB-D saliency detection models, \eg~UCNet \cite{ucnet_sal} (62M), BBSNet \cite{fan2020bbsnet} (49M), while smaller than JL-DCF \cite{fu2020jl} (144M). The extra parameters mainly come from our separate encoder with ResNet50 backbone for auxiliary depth estimation.
At test time, it cost 0.2 second to process each image of size $352\times352$, which is comparable with benchmark techniques.

\subsection{Ablation Study}
We thoroughly analyse the proposed solutions with extra experiments, and show results in Table \ref{tab:ablation_study}. Note that, all the following experiments are trained with ResNet50 backbone with training/testing image size of 352.

Firstly, we train an RGB COD model with only the RGB image, and the performance is shown as \enquote{Base}. The network structure of \enquote{Base} is the same as our auxiliary depth estimation network, except that we use structure-aware loss function \cite{wei2020f3net} for \enquote{Base}. Then, we introduce our auxiliary depth estimation branch to \enquote{Base}, and obtain \enquote{ADE}, where there exists no RGB and depth feature interaction. We further add the \enquote{Multi-modal Fusion} module to \enquote{ADE} to produce extra RGB-D camouflage map,
and show performance as \enquote{A\_D}. Note that, until now, all the discussed models are deterministic.
We intend to explore the depth contribution with a generative adversarial network, and we then add latent variable $z$ to \enquote{A\_D} with an extra discriminator
to obtain a generative adversarial network based RGB-D COD network, which is our final model, shown as \enquote{Ours}.

\begin{table}[t!]
  \centering
  \footnotesize
  \renewcommand{\arraystretch}{1.1}
  \renewcommand{\tabcolsep}{0.55mm}
  \caption{Performance of ablation study models.}
  \begin{tabular}{l|cc|cc|cc|cc}
  \hline
  &\multicolumn{2}{c|}{CAMO~\cite{le2019anabranch}}&\multicolumn{2}{c|}{CHAMELEON~\cite{Chameleon2018}}&\multicolumn{2}{c|}{COD10K~\cite{fan2020camouflaged}}&\multicolumn{2}{c}{NC4K~\cite{yunqiu_cod21}} \\
    Method &$F_{\beta}\uparrow$&$\mathcal{M}\downarrow$&$F_{\beta}\uparrow$&$\mathcal{M}\downarrow$&$F_{\beta}\uparrow$&$\mathcal{M}\downarrow$&$F_{\beta}\uparrow$&$\mathcal{M}\downarrow$ \\
  \hline
  Base &.751 & .079 &.815 & .037 &.696 & .039 &.789 & .050 \\ 
ADE &.735 & .084 &.826 & .033 &.699 & .039 &.794 & .050 \\
A\_D  &.746 & .079 &.835 & .032 &.702 & .038 &.793 & .049 \\ \hline
Ours &.759 & .077 &.836 & .032 &.705 & .037 &.798 & .049\\
   \hline
  \end{tabular}
  \label{tab:ablation_study}
\end{table}

We observe competing performance of \enquote{Base} compared with the state-of-the-art camouflaged object detection networks, indicating effectiveness of the simple base model. As there exists no interaction of RGB feature and depth feature, the performance of \enquote{ADE} is similar to \enquote{Base}. At the same time, as the weight of each task is quite important for a multi-task learning framework (within \enquote{ADE}, we achieve simultaneous camouflaged object detection and monocular depth estimation), the performance of \enquote{ADE} can be further tuned with more effective task-relationship models \cite{kendall2018multi}.
Build upon \enquote{ADE} with extra \enquote{Multi-modal Fusion} block, we notice improved performance of \enquote{A\_D} compared with \enquote{ADE}, which explains effectiveness of our \enquote{Multi-modal Fusion} block. Then, we add extra latent variable $z$ and a fully convolutional discriminator to \enquote{A\_D}, and perform multi-modal confidence-aware learning.
The improved performance of \enquote{Ours}
explains effectiveness of our depth contribution exploration solutions.

\begin{figure}[htbp]
   \begin{center}
   \begin{tabular}{{c@{ } c@{ } c@{ } c@{ } c@{ } c@{ } }}
    {\includegraphics[width=0.15\linewidth]{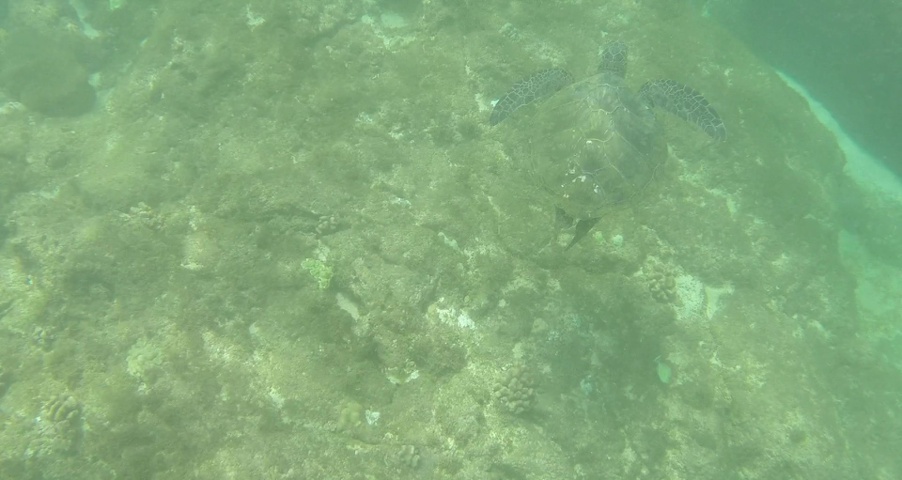}}&
    {\includegraphics[width=0.15\linewidth]{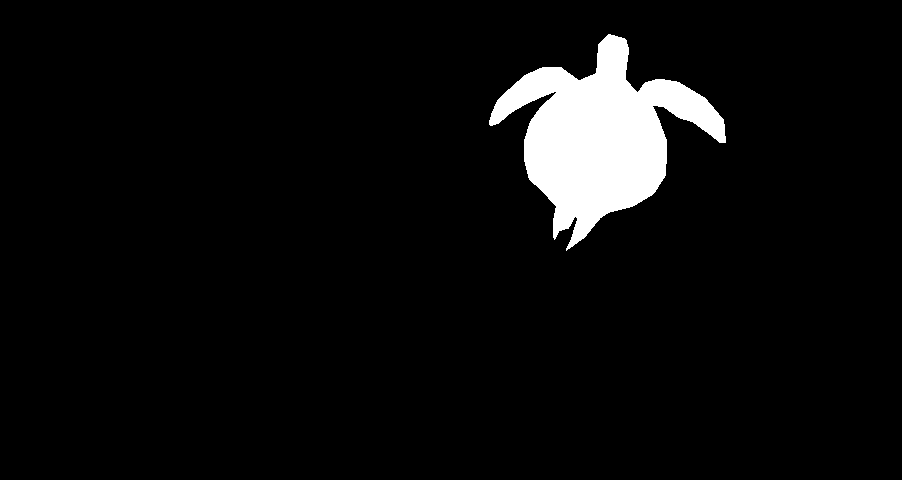}}&
    {\animategraphics[width=0.15\linewidth,autoplay, loop]{4}{figure_multi_res/ani/cod_rgb_00135_}{0}{15}} &
    {\includegraphics[width=0.15\linewidth]{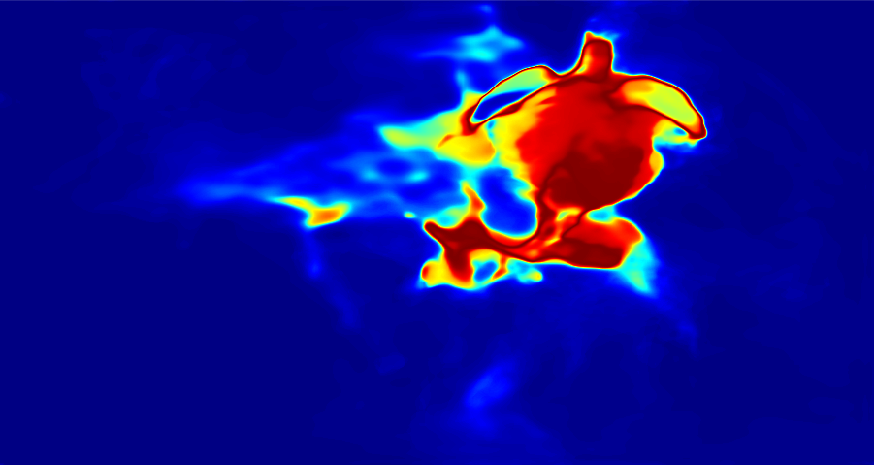}}&
    {\animategraphics[width=0.15\linewidth,autoplay, loop]{4}{figure_multi_res/ani/cod_rgbd_00135_}{0}{15}} &
    {\includegraphics[width=0.15\linewidth]{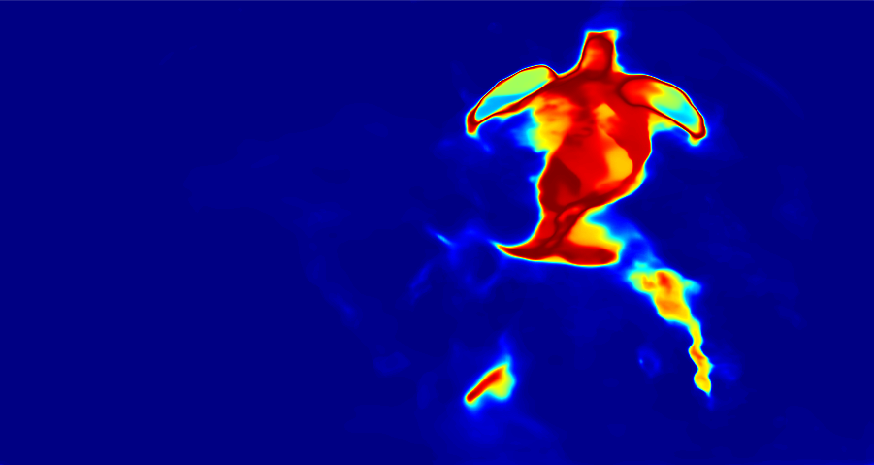}} \\
    
    {\includegraphics[width=0.15\linewidth]{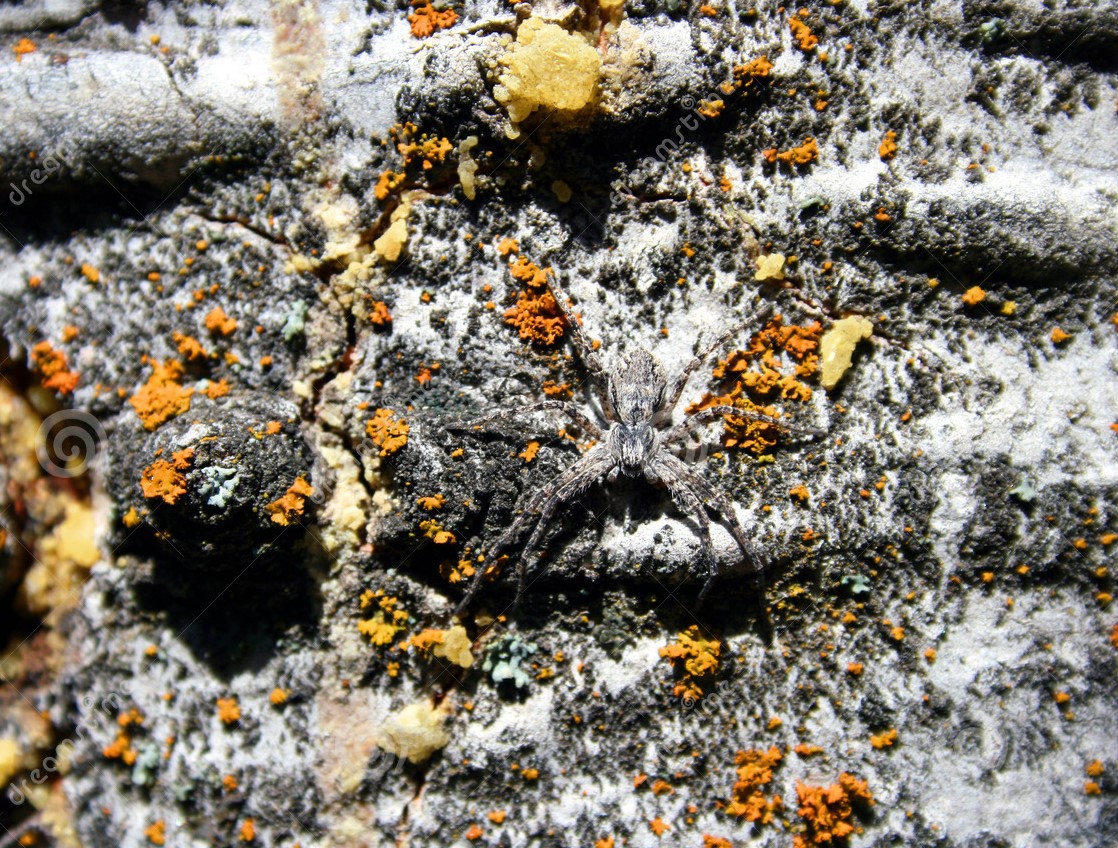}}&
    {\includegraphics[width=0.15\linewidth]{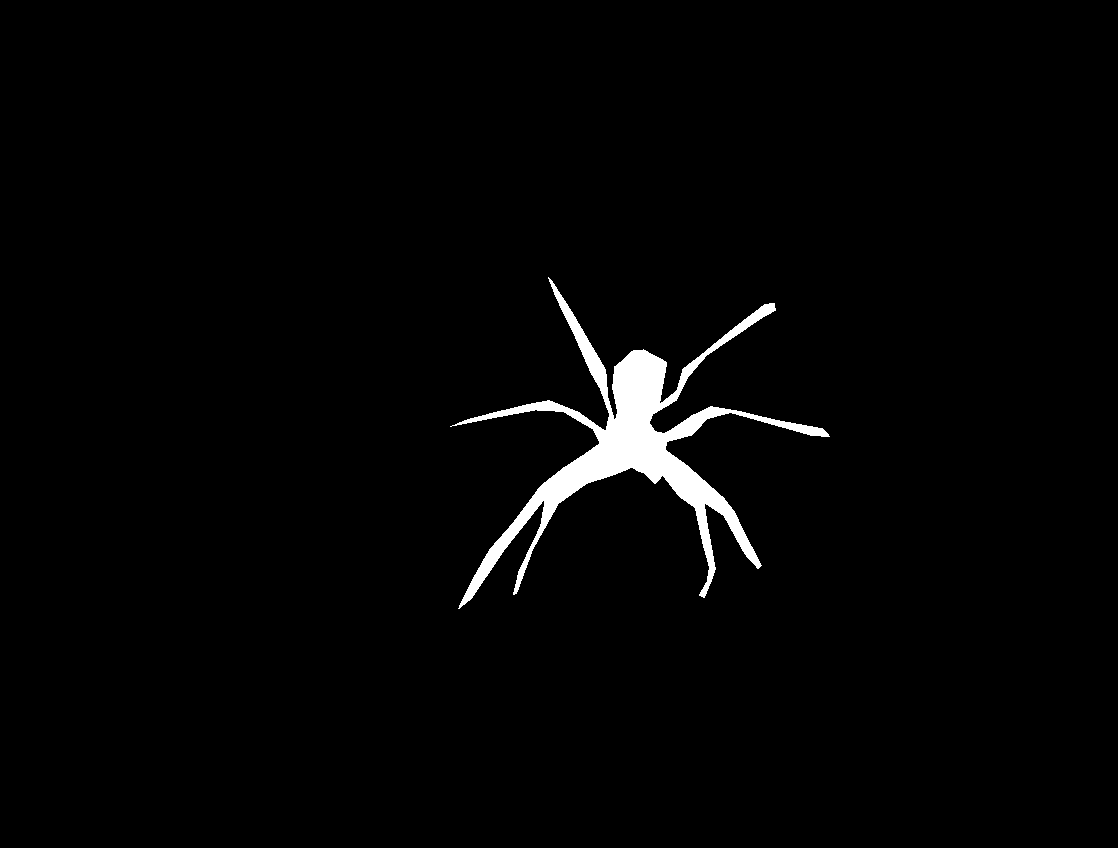}}&
    {\animategraphics[width=0.15\linewidth,autoplay, loop]{4}{figure_multi_res/ani/cod_rgb_00500_}{0}{15}} &
    {\includegraphics[width=0.15\linewidth]{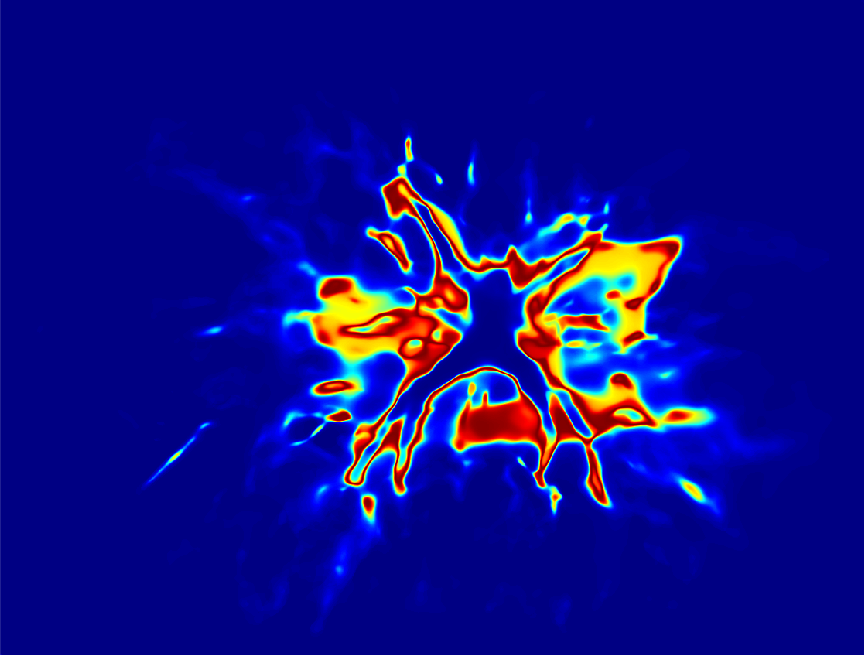}}&
    {\animategraphics[width=0.15\linewidth,autoplay, loop]{4}{figure_multi_res/ani/cod_rgbd_00500_}{0}{15}} &
    {\includegraphics[width=0.15\linewidth]{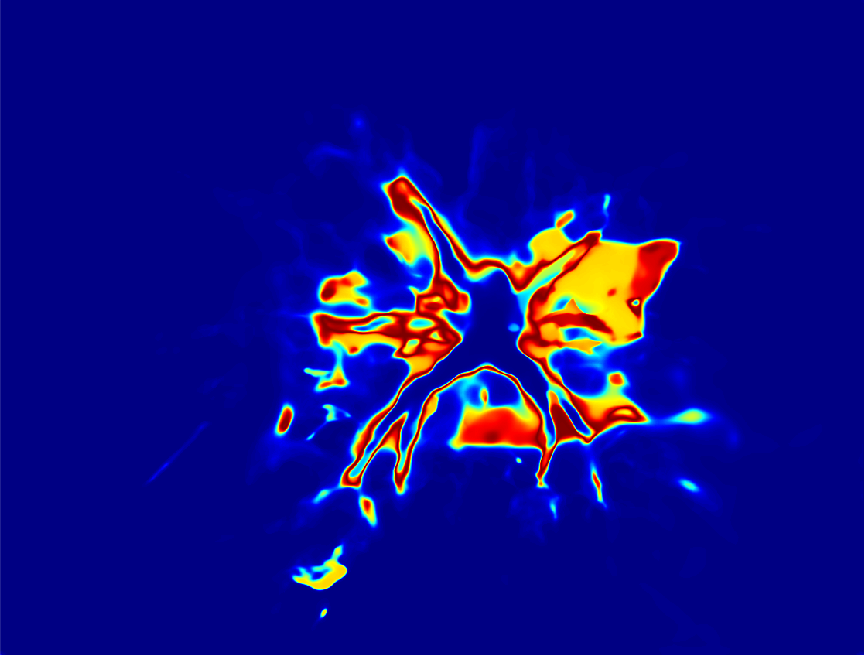}} 
    \\
    \footnotesize{Image} 
    &\footnotesize{GT} 
    &\scriptsize{Ours\_RGB}
    &\scriptsize{$U_{rgb}$}
    &\scriptsize{Ours\_RGBD}
    &\scriptsize{$U_{rgbd}$} \\
   \end{tabular}
   \end{center}
    \caption{Predictive uncertainties deduced from predictions with different input sizes.
    } 
    \label{fig:ani_pu}
\end{figure}

\subsection{Training/Testing size and Backbone Analysis}
Training/testing image size is usually an important factor for dense prediction tasks, and we observe it's more important for camouflaged object detection, where higher resolution training images can always lead to better performance. In Table \ref{tab:benchmark_model_comparison_rgb}, we summarize the training/testing image sizes of existing methods, and train our model with different image sizes.
We observe consistently improved performance with larger image sizes. The main reason is that lower resolution can be seen as adversarial attack for camouflaged object detection. Camouflaged objects usually share a similar appearance as the background, and it's more difficult to discover camouflaged objects in a small size image than in a large size image.
In this way, reducing the images sizes of a training dataset can be seen as improving the difficulty of the task, and training models with too many hard samples will limit its generalization ability.

We further notice that testing with different sizes also affects model predictions. Models perform best when we have the same training and testing data sizes.
By associating predictions of a single input with a series of sizes, we can deduce the predictive uncertainty that lies intrinsically in the model and the task themselves (see Fig. \ref{fig:ani_pu}).

Further,
we notice backbone of SINet-V2 \cite{fan2021concealed} is Res2Net50 \cite{res2net}. For fair comparison, and also to investigate model performance \wrt~backbone networks, we train extra model with Res2Net50 backbone. The consistent better performance of our model with Res2Net50 \cite{res2net} backbone compared with SINet-V2 \cite{fan2021concealed} validate our solution.

\begin{table}[t!]
  \centering
  \footnotesize
  \renewcommand{\arraystretch}{1.2}
  \renewcommand{\tabcolsep}{0.7mm}
  \caption{Performance of RGB-D COD with existing sensor depth based multi-modal learning framework.}
  \begin{tabular}{l|cc|cc|cc|cc}
  \hline
  &\multicolumn{2}{c|}{CAMO~\cite{le2019anabranch}}&\multicolumn{2}{c|}{CHAMELEON~\cite{Chameleon2018}}&\multicolumn{2}{c|}{COD10K~\cite{fan2020camouflaged}}&\multicolumn{2}{c}{NC4K~\cite{yunqiu_cod21}} \\
    Method &$F_{\beta}\uparrow$&$\mathcal{M}\downarrow$&$F_{\beta}\uparrow$&$\mathcal{M}\downarrow$&$F_{\beta}\uparrow$&$\mathcal{M}\downarrow$&$F_{\beta}\uparrow$&$\mathcal{M}\downarrow$  \\
  \hline
  Base  &.751 & .079 &.815 & .037 &.696 & .039 &.789 & .050 \\ 
Early &.718 & .092 &.816 & .037 &.677 & .042 &.770 & .056 \\
Cross  &.680 & .102 &.826 & .035 &.694 & .040 &.772 & .060 \\
Late &.689 & .100 &.808 & .043 &.681 & .044 &.752 & .068\\ 
   \hline
  \end{tabular}
  \label{tab:multi_modal_sensor_depth}
\end{table}

\begin{table}[!ht]
  \centering
  \footnotesize
  \renewcommand{\arraystretch}{1.2}
  \renewcommand{\tabcolsep}{0.5mm}
  \caption{Multi-modal learning performance for sensor depth.
  }
  \begin{tabular}{l|cc|cc|cc|cc|cc}
  \hline
  &\multicolumn{2}{c|}{NJU2K~\cite{NJU2000}}&\multicolumn{2}{c|}{SSB~\cite{niu2012leveraging}}&\multicolumn{2}{c|}{DES~\cite{cheng2014depth}}&\multicolumn{2}{c|}{NLPR~\cite{peng2014rgbd}}&\multicolumn{2}{c}{SIP~\cite{sip_dataset}} \\
    Method &$F_{\beta}\uparrow$&$\mathcal{M}\downarrow$&$F_{\beta}\uparrow$&$\mathcal{M}\downarrow$&$F_{\beta}\uparrow$&$\mathcal{M}\downarrow$&$F_{\beta}\uparrow$&$\mathcal{M}\downarrow$&$F_{\beta}\uparrow$&$\mathcal{M}\downarrow$ \\ \hline
   Base &.898 &.036 &.884 &.037  &.904 &.022  &.897 &.024  &.866 &.049  \\
   Early &.904 &.036 &.887 &.037  &.924 &.017  &.904 &.022 &.882 &.047  \\
   Cross &.905 &.036 &.890 &.036  &.928 &.016  &.904 &.024 &.886 &.047  \\
   Late &.897 &.040 &.863 &.053  &.906 &.020 &.897 &.024 &.886 &.048  \\
   \hline 
  \end{tabular}
  \label{tab:ablation_rgbd_sod}
\end{table}

\subsection{Multi-Modal Learning Discussion}
We investigate the \enquote{sensor depth} based multi-modal learning strategies on
our \enquote{generated depth} scenario.

\noindent\textbf{Sensor depth based multi-modal learning framework for COD:} With depth from monocular depth estimation method, following conventional multi-modal learning pipeline,
we can train directly the RGB-D COD task.
Specifically, we introduce three different settings, namely early fusion model (\enquote{Early}), cross-level fusion model (\enquote{Cross}) and late fusion model (\enquote{Late}), and show model performance in Table \ref{tab:multi_modal_sensor_depth}. For \enquote{Early}, we concatenate $x$ and $d$ at the input layer, and feed it to a $3\times3$ convolutional layers to obtain a feature map of channel size 3, which is then
fed
to the RGB COD network to produce camouflage prediction (without the latent variable $z$ as input). For \enquote{Cross}, we re-purpose the auxiliary depth estimation to depth based COD, and
keep the \enquote{Multi-modal Fusion} module. In this way, we have camouflage prediction from the depth branch, the RGB branch and also the RGB-D branch, and prediction of the RGB-D branch is used for performance evaluation.
For \enquote{Late}, we have two copies of RGB COD network that take RGB and depth as input to produce two copies of predictions. We concatenate the two predictions and feed it to a $3\times3$ convolutional layers to produce our final prediction. Table \ref{tab:multi_modal_sensor_depth} shows that the existing \enquote{sensor depth} based multi-modal learning framework works poorly on our \enquote{generated depth} scenario, leading to inferior performance compared to RGB image based \enquote{Base} model.

\noindent\textbf{Effectiveness of the multi-modal learning framework in Table \ref{tab:multi_modal_sensor_depth} with \enquote{sensor depth}:}
Similar to camouflaged object detection, salient object detection \cite{wei2020f3net} is also defined as class-agnostic binary segmentation task. We then analyse how the sensor depth based multi-modal learning framework performs for RGB-D salient object detection \cite{fan2020bbsnet}, and show performance in Table \ref{tab:ablation_rgbd_sod}, where the depth are the raw \enquote{sensor depth} instead of the \enquote{generated depth}. Table \ref{tab:ablation_rgbd_sod} shows that the multi-modal learning framework in Table \ref{tab:ablation_rgbd_sod} performs well with sensor depth scenario, and the different conclusion with \enquote{sensor depth} and \enquote{generated depth} further explain the value of our work.

\subsection{When is the Generated Depth More Helpful?}
\begin{figure}[htbp]
   \begin{center}
   \begin{tabular}{{c@{ } c@{ } c@{ } c@{ } c@{ } c@{ } }}
    {\includegraphics[width=0.15\linewidth]{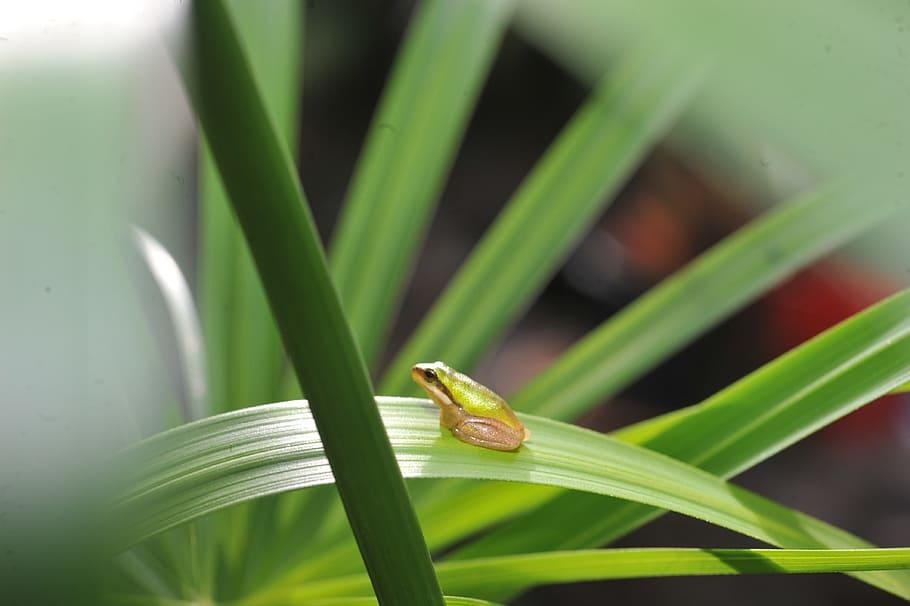}}&
    {\includegraphics[width=0.15\linewidth]{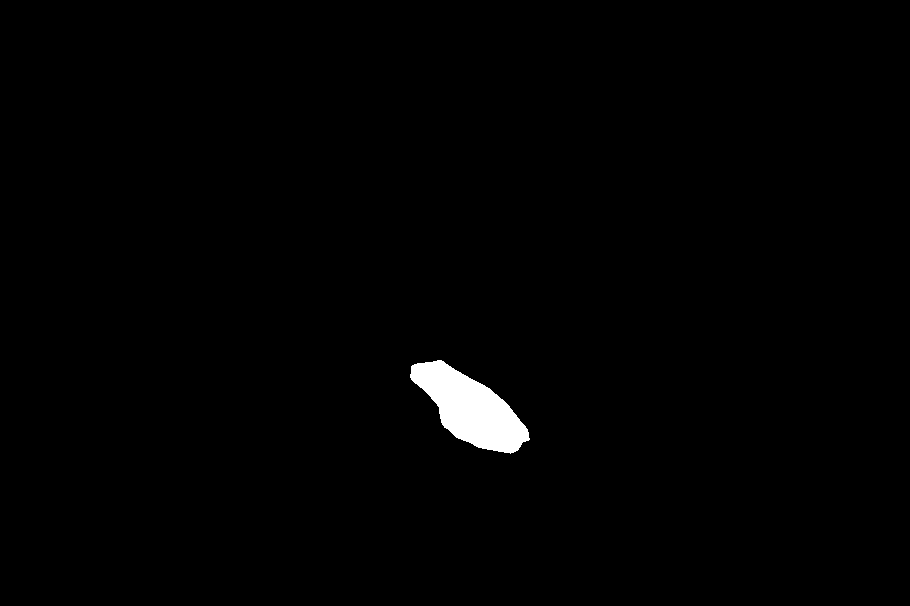}}&
    {\includegraphics[width=0.15\linewidth]{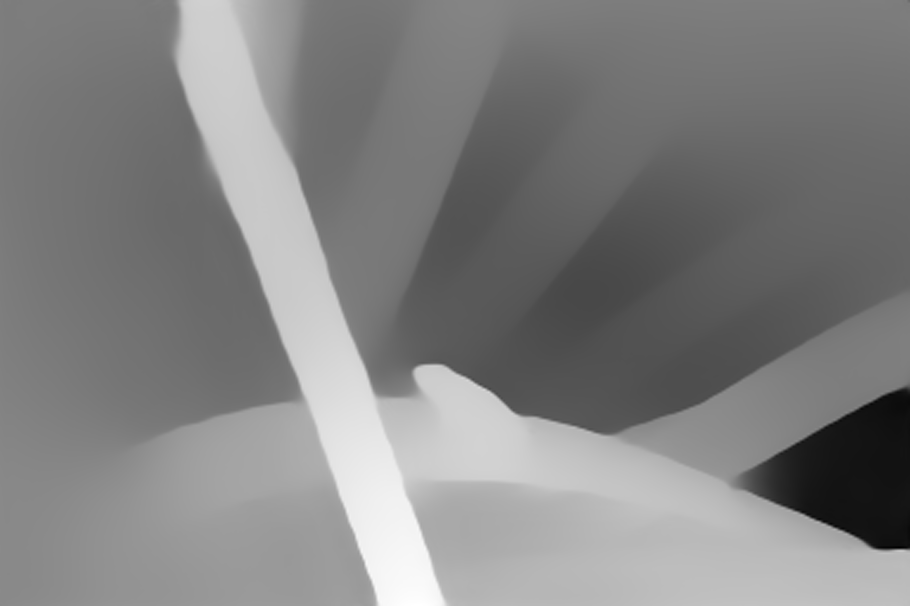}}&
    {\includegraphics[width=0.15\linewidth]{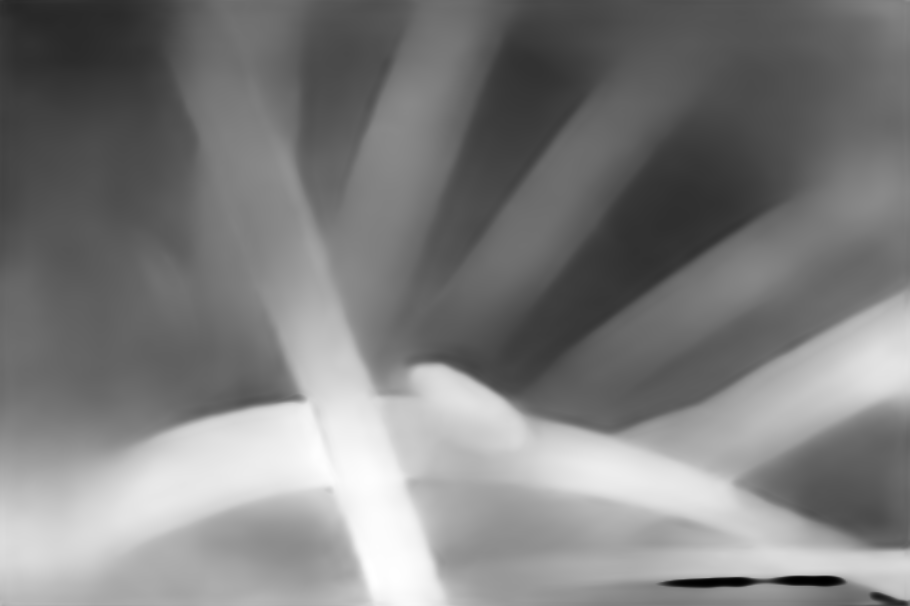}}&
    {\includegraphics[width=0.15\linewidth]{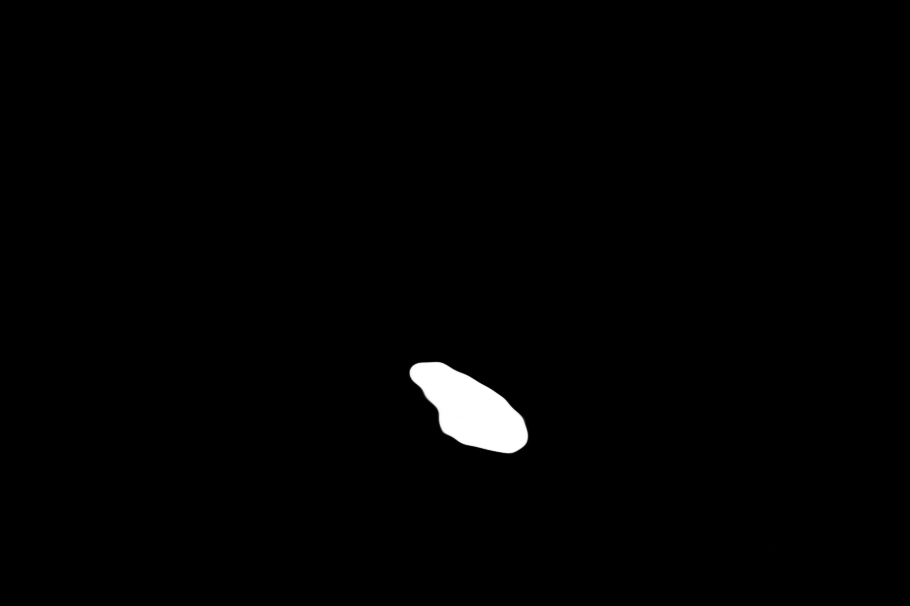}}&
    {\includegraphics[width=0.15\linewidth]{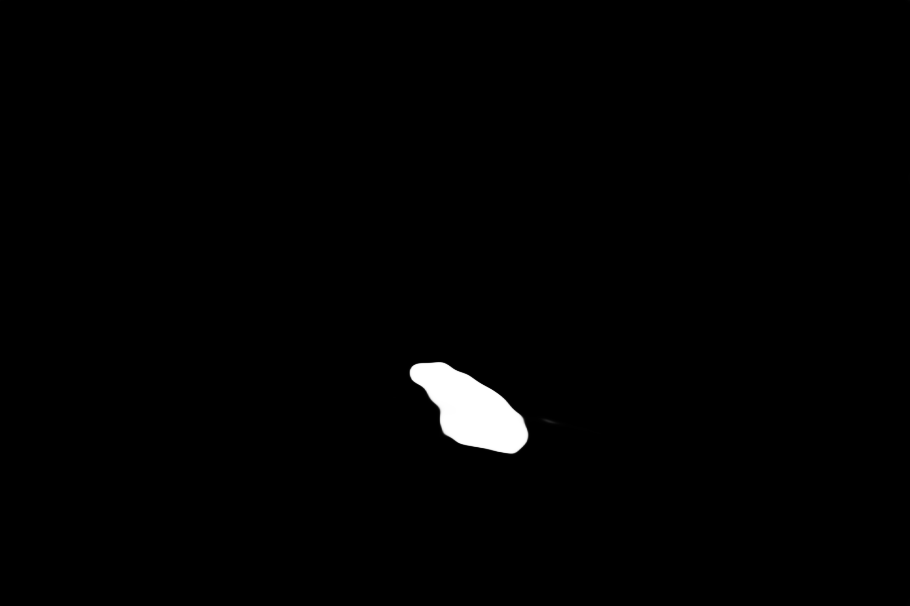}}
    \\
    {\includegraphics[width=0.15\linewidth]{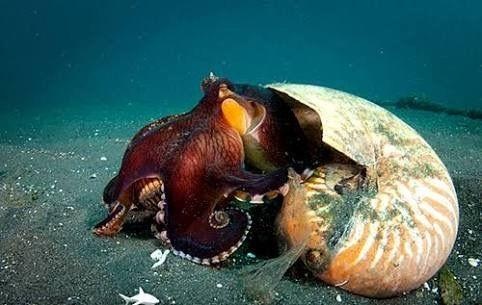}}&
    {\includegraphics[width=0.15\linewidth]{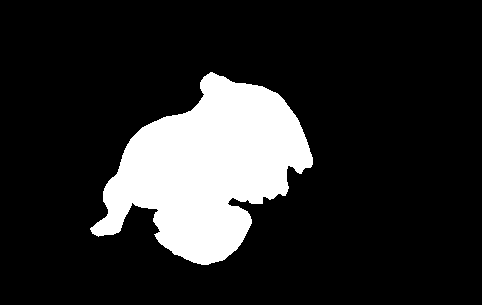}}&
    {\includegraphics[width=0.15\linewidth]{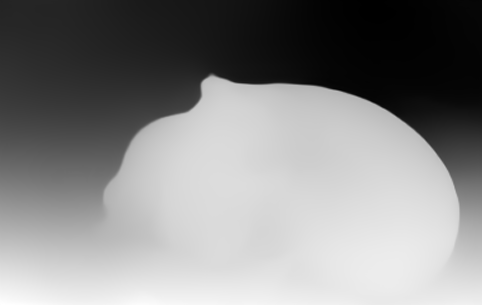}}&
    {\includegraphics[width=0.15\linewidth]{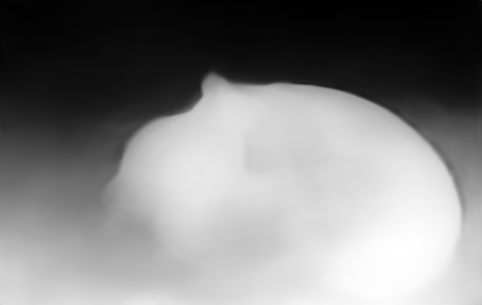}}&
    {\includegraphics[width=0.15\linewidth]{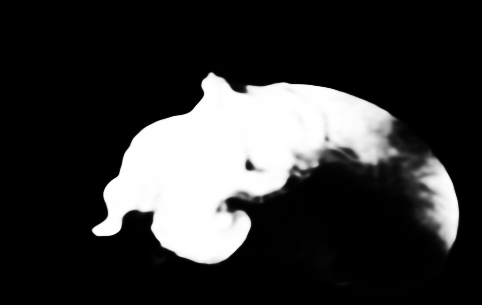}}&
    {\includegraphics[width=0.15\linewidth]{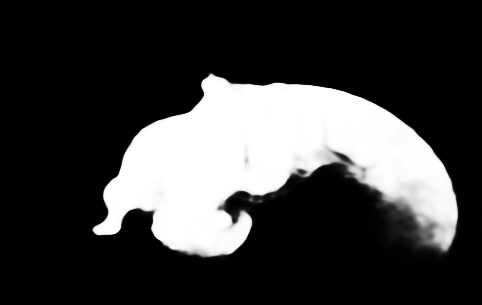}}
    \\
    \footnotesize{Image} 
    &\footnotesize{GT} 
    &\footnotesize{Depth}
    &\footnotesize{$d^{'}$}
    &\footnotesize{Ours\_{RGB}}
    &\footnotesize{Ours\_{RGBD}}\\
   \end{tabular}
   \end{center}
    \caption{Scenarios when the RGB COD branch outperforms RGB-D COD branch.
    } 
    \label{fig:depth_help}
\end{figure}

RGB image captures appearance information of a scenario, while depth captures geometric information, indicating the distance of objects to the camera. It's usually claimed that extra data from different modal can lead to better understanding of the same scene. In this paper, we investigate how the generated depth can improve camouflaged object detection.
Although the overall performance indicates that depth can be helpful if we use it properly, we still notice samples where the final camouflage map from the RGB-D branch is inferior to that from the RGB branch as shown in Fig.~\ref{fig:depth_help}.

Fig.~\ref{fig:depth_help} shows that depth can be a double-edged sword.
In good cases, both RGB and RGB-D COD branches provide decent segmentation results, and we
notice that depth predictions ($d'$) in these scenarios tend to highlight the camouflaged objects better compared with the
original generated depths ($d$).
Further, when we can observe the camouflaged objects from the generated depth ($d$),
the depth $d$ can always benefit the RGB-D COD.
In hard cases, especially when camouflaged objects dominate the view and have large depth variation, RGB-D COD prediction can be misled by the depth estimation task.

\section{Conclusions}
We tackle the problem of camouflaged object detection (COD) with the aid of depth, which is generated with the existing monocular depth estimation method \cite{MiDaS_Ranftl_2020_TPAMI}. To effectively explore depth contribution, we introduce an auxiliary depth estimation module, and a probabilistic learning network via a generative adversarial network, where the modal related confidence is estimated and serves as a weight for multi-modal confidence-aware learning. With extensive experimental results, we conclude that the existing \enquote{sensor depth} based multi-modal learning pipelines perform poorly with the \enquote{generated depth} setting, leading to inferior performance compared to the base model with only RGB image used. Instead, we suggest performing auxiliary depth estimation with an effective depth confidence estimation module to prevent the model from being dominated by the noisy depth, which has been proven effective in our \enquote{generated depth} based RGB-D COD task.


{\small
\bibliographystyle{ieee_fullname}
\bibliography{RGBD_COD_Reference}

\begin{thebibliography}{10}\itemsep=-1pt

\bibitem{adams2019disruptive}
Wendy~J Adams, Erich~W Graf, and Matt Anderson.
\newblock Disruptive coloration and binocular disparity: breaking camouflage.
\newblock {\em Proceedings of the Royal Society B}, 286(1896):20182045, 2019.

\bibitem{bhajantri2006camouflage}
Nagappa~U Bhajantri and P Nagabhushan.
\newblock Camouflage defect identification: a novel approach.
\newblock In {\em International Conference on Information Technology}, pages
  145--148, 2006.

\bibitem{chen2018progressively}
Hao Chen and Youfu Li.
\newblock Progressively complementarity-aware fusion network for {RGB-D}
  salient object detection.
\newblock In {\em IEEE Conference on Computer Vision and Pattern Recognition
  (CVPR)}, pages 3051--3060, 2018.

\bibitem{chen2017deeplab}
Liang-Chieh Chen, George Papandreou, Iasonas Kokkinos, Kevin Murphy, and Alan~L
  Yuille.
\newblock Deeplab: Semantic image segmentation with deep convolutional nets,
  atrous convolution, and fully connected {CRF}s.
\newblock {\em IEEE Transactions on Pattern Analysis and Machine Intelligence
  (TPAMI)}, 40(4):834--848, 2017.

\bibitem{cheng2014depth}
Yupeng Cheng, Huazhu Fu, Xingxing Wei, Jiangjian Xiao, and Xiaochun Cao.
\newblock Depth enhanced saliency detection method.
\newblock In {\em Proceedings of international conference on internet
  multimedia computing and service}, pages 23--27, 2014.

\bibitem{copeland1997models}
Anthony~C Copeland and Mohan~M Trivedi.
\newblock Models and metrics for signature strength evaluation of camouflaged
  targets.
\newblock In {\em Algorithms for Synthetic Aperture Radar Imagery IV}, volume
  3070, pages 194--199, 1997.

\bibitem{deng2015semantic}
Zhuo Deng, Sinisa Todorovic, and Longin Jan~Latecki.
\newblock Semantic segmentation of {RGBD} images with mutex constraints.
\newblock In {\em IEEE Conference on Computer Vision and Pattern Recognition
  (CVPR)}, pages 1733--1741, 2015.

\bibitem{dong2021towards}
Bo Dong, Mingchen Zhuge, Yongxiong Wang, Hongbo Bi, and Geng Chen.
\newblock Towards accurate camouflaged object detection with mixture
  convolution and interactive fusion.
\newblock {\em arXiv preprint arXiv:2101.05687}, 2021.

\bibitem{fan2017structure}
Deng-Ping Fan, Ming-Ming Cheng, Yun Liu, Tao Li, and Ali Borji.
\newblock Structure-measure: A new way to evaluate foreground maps.
\newblock In {\em IEEE International Conference on Computer Vision (ICCV)},
  pages 4548--4557, 2017.

\bibitem{fan2018}
Deng-Ping Fan, Cheng Gong, Yang Cao, Bo Ren, Ming-Ming Cheng, and Ali Borji.
\newblock Enhanced-alignment measure for binary foreground map evaluation.
\newblock In {\em International Joint Conference on Artificial Intelligence
  (IJCAI)}, pages 698--704, 2018.

\bibitem{fan2021concealed}
Deng-Ping Fan, Ge-Peng Ji, Ming-Ming Cheng, and Ling Shao.
\newblock Concealed object detection.
\newblock {\em IEEE Transactions on Pattern Analysis and Machine Intelligence
  (TPAMI)}, 2021.

\bibitem{fan2020camouflaged}
Deng-Ping Fan, Ge-Peng Ji, Guolei Sun, Ming-Ming Cheng, Jianbing Shen, and Ling
  Shao.
\newblock Camouflaged object detection.
\newblock In {\em IEEE Conference on Computer Vision and Pattern Recognition
  (CVPR)}, pages 2777--2787, 2020.

\bibitem{fan2020pranet}
Deng-Ping Fan, Ge-Peng Ji, Tao Zhou, Geng Chen, Huazhu Fu, Jianbing Shen, and
  Ling Shao.
\newblock {PraNet}: Parallel reverse attention network for polyp segmentation.
\newblock In {\em International Conference on Medical Image Computing and
  Computer-Assisted Intervention}, pages 263--273, 2020.

\bibitem{sip_dataset}
Deng-Ping Fan, Zheng Lin, Zhao Zhang, Menglong Zhu, and Ming-Ming Cheng.
\newblock {Rethinking RGB-D salient object detection: models, datasets, and
  large-scale benchmarks}.
\newblock {\em IEEE Transactions on Neural Networks and Learning Systems
  (TNNLS)}, 2020.

\bibitem{fan2020bbsnet}
Deng-Ping Fan, Yingjie Zhai, Ali Borji, Jufeng Yang, and Ling Shao.
\newblock {BBS-Net}: {RGB-D} salient object detection with a bifurcated
  backbone strategy network.
\newblock In {\em European Conference on Computer Vision (ECCV)}, pages
  275--292, 2020.

\bibitem{fang2019camouflage}
Zheng Fang, Xiongwei Zhang, Xiaotong Deng, Tieyong Cao, and Changyan Zheng.
\newblock Camouflage people detection via strong semantic dilation network.
\newblock In {\em Proceedings of the ACM Turing Celebration Conference-China},
  pages 1--7, 2019.

\bibitem{farabet2012learning}
Clement Farabet, Camille Couprie, Laurent Najman, and Yann LeCun.
\newblock Learning hierarchical features for scene labeling.
\newblock {\em IEEE Transactions on Pattern Analysis and Machine Intelligence
  (TPAMI)}, 35(8):1915--1929, 2012.

\bibitem{feng2016local}
David Feng, Nick Barnes, Shaodi You, and Chris McCarthy.
\newblock Local background enclosure for {RGB-D} salient object detection.
\newblock In {\em IEEE Conference on Computer Vision and Pattern Recognition
  (CVPR)}, pages 2343--2350, 2016.

\bibitem{feng2015camouflage}
Xue Feng, Cui Guoying, Hong Richang, and Gu Jing.
\newblock Camouflage texture evaluation using a saliency map.
\newblock {\em Multimedia Systems}, 21(2):169--175, 2015.

\bibitem{fu2020jl}
Keren Fu, Deng-Ping Fan, Ge-Peng Ji, and Qijun Zhao.
\newblock {JL-DCF}: Joint learning and densely-cooperative fusion framework for
  {RGB-D} salient object detection.
\newblock In {\em IEEE Conference on Computer Vision and Pattern Recognition
  (CVPR)}, pages 3052--3062, 2020.

\bibitem{res2net}
S. Gao, M. Cheng, K. Zhao, X. Zhang, M. Yang, and P. Torr.
\newblock Res2net: A new multi-scale backbone architecture.
\newblock {\em IEEE Transactions on Pattern Analysis and Machine Intelligence
  (TPAMI)}, 43(02):652--662, 2021.

\bibitem{georgecvpr2021}
Anjith George and Sebastien Marcel.
\newblock Cross modal focal loss for {RGBD} face anti-spoofing.
\newblock In {\em IEEE Conference on Computer Vision and Pattern Recognition
  (CVPR)}, pages 7882--7891, 2021.

\bibitem{ssim_depth}
Cl{\'{e}}ment Godard, Oisin {Mac Aodha}, and Gabriel~J. Brostow.
\newblock Unsupervised monocular depth estimation with left-right consistency.
\newblock In {\em IEEE Conference on Computer Vision and Pattern Recognition
  (CVPR)}, 2017.

\bibitem{monodepth2_Clement_2019_ICCV}
Cl{\'{e}}ment Godard, Oisin {Mac Aodha}, Michael Firman, and Gabriel~J.
  Brostow.
\newblock Digging into self-supervised monocular depth prediction.
\newblock In {\em IEEE International Conference on Computer Vision (ICCV)},
  pages 3828--3838, 2019.

\bibitem{gan_raw}
Ian Goodfellow, Jean Pouget-Abadie, Mehdi Mirza, Bing Xu, David Warde-Farley,
  Sherjil Ozair, Aaron Courville, and Yoshua Bengio.
\newblock Generative adversarial nets.
\newblock In {\em Conference on Neural Information Processing Systems
  (NeurIPS)}, pages 2672--2680, 2014.

\bibitem{on_calibration}
Chuan Guo, Geoff Pleiss, Yu Sun, and Kilian~Q. Weinberger.
\newblock On calibration of modern neural networks.
\newblock In {\em International Conference on Machine Learning (ICML)}, pages
  1321--1330, 2017.

\bibitem{gupta2013perceptual}
Saurabh Gupta, Pablo Arbelaez, and Jitendra Malik.
\newblock Perceptual organization and recognition of indoor scenes from {RGB-D}
  images.
\newblock In {\em IEEE Conference on Computer Vision and Pattern Recognition
  (CVPR)}, pages 564--571, 2013.

\bibitem{hall2020platform}
Joanna~R Hall, Olivia Matthews, Timothy~N Volonakis, Eric Liggins, Karl~P
  Lymer, Roland Baddeley, Innes~C Cuthill, and Nicholas~E Scott-Samuel.
\newblock A platform for initial testing of multiple camouflage patterns.
\newblock {\em Defence Technology}, 2020.

\bibitem{he2017mask}
Kaiming He, Georgia Gkioxari, Piotr Doll{\'a}r, and Ross Girshick.
\newblock Mask {R-CNN}.
\newblock In {\em IEEE International Conference on Computer Vision (ICCV)},
  pages 2961--2969, 2017.

\bibitem{he2016deep}
Kaiming He, Xiangyu Zhang, Shaoqing Ren, and Jian Sun.
\newblock Deep residual learning for image recognition.
\newblock In {\em IEEE Conference on Computer Vision and Pattern Recognition
  (CVPR)}, pages 770--778, 2016.

\bibitem{resnet_he}
Kaiming He, Xiangyu Zhang, Shaoqing Ren, and Jian Sun.
\newblock Deep residual learning for image recognition.
\newblock In {\em 2016 IEEE Conference on Computer Vision and Pattern
  Recognition (CVPR)}, pages 770--778, 2016.

\bibitem{hu2019acnet}
Xinxin Hu, Kailun Yang, Lei Fei, and Kaiwei Wang.
\newblock {ACNet}: Attention based network to exploit complementary features
  for {RGBD} semantic segmentation.
\newblock In {\em IEEE International Conference on Image Processing (ICIP)},
  pages 1440--1444, 2019.

\bibitem{hung2018adversarial}
Wei-Chih Hung, Yi-Hsuan Tsai, Yan-Ting Liou, Yen-Yu Lin, and Ming-Hsuan Yang.
\newblock Adversarial learning for semi-supervised semantic segmentation.
\newblock {\em arXiv preprint arXiv:1802.07934}, 2018.

\bibitem{NJU2000}
Ran Ju, Yang Liu, Tongwei Ren, Ling Ge, and Gangshan Wu.
\newblock Depth-aware salient object detection using anisotropic
  center-surround difference.
\newblock {\em Signal Processing: Image Communication}, 38:115--126, 2015.

\bibitem{kelman2008review}
Emma~J Kelman, Daniel Osorio, and Roland~J Baddeley.
\newblock A review of cuttlefish camouflage and object recognition and evidence
  for depth perception.
\newblock {\em Journal of Experimental Biology}, 211(11):1757--1763, 2008.

\bibitem{what_uncertainty}
Alex Kendall and Yarin Gal.
\newblock What uncertainties do we need in bayesian deep learning for computer
  vision?
\newblock In {\em Conference on Neural Information Processing Systems
  (NeurIPS)}, pages 5580--5590, 2017.

\bibitem{kendall2018multi}
Alex Kendall, Yarin Gal, and Roberto Cipolla.
\newblock Multi-task learning using uncertainty to weigh losses for scene
  geometry and semantics.
\newblock In {\em IEEE Conference on Computer Vision and Pattern Recognition
  (CVPR)}, pages 7482--7491, 2018.

\bibitem{kong2020sde}
Lingkai Kong, Jimeng Sun, and Chao Zhang.
\newblock {SDE-Net}: Equipping deep neural networks with uncertainty estimates.
\newblock {\em International Conference on Machine Learning (ICML)}, pages
  5405--5415, 2020.

\bibitem{simple_scalable_uncertainty}
Balaji Lakshminarayanan, Alexander Pritzel, and Charles Blundell.
\newblock Simple and scalable predictive uncertainty estimation using deep
  ensembles.
\newblock In {\em Conference on Neural Information Processing Systems
  (NeurIPS)}, pages 6402--6413, 2017.

\bibitem{le2019anabranch}
Trung-Nghia Le, Tam~V Nguyen, Zhongliang Nie, Minh-Triet Tran, and Akihiro
  Sugimoto.
\newblock Anabranch network for camouflaged object segmentation.
\newblock {\em Computer Vision and Image Understanding (CVIU)}, 184:45--56,
  2019.

\bibitem{lev2004plant}
Simcha Lev-Yadun, Amots Dafni, Moshe~A Flaishman, Moshe Inbar, Ido Izhaki, Gadi
  Katzir, and Gidi Ne'eman.
\newblock Plant coloration undermines herbivorous insect camouflage.
\newblock {\em BioEssays}, 26(10):1126--1130, 2004.

\bibitem{fusion_camo}
Shuai. {Li}, Dinei. {Florencio}, Wanqing {Li}, Yaqin {Zhao}, and Chris {Cook}.
\newblock A fusion framework for camouflaged moving foreground detection in the
  wavelet domain.
\newblock {\em IEEE Transactions on Image Processing (TIP)}, 27(8):3918--3930,
  2018.

\bibitem{FrozenPeople_li_2019_CVPR}
Zhengqi Li, Tali Dekel, Forrester Cole, Richard Tucker, Noah Snavely, Ce Liu,
  and William~T Freeman.
\newblock Learning the depths of moving people by watching frozen people.
\newblock In {\em IEEE Conference on Computer Vision and Pattern Recognition
  (CVPR)}, pages 4521--4530, 2019.

\bibitem{lin2017cascaded}
Di Lin, Guangyong Chen, Daniel Cohen-Or, Pheng-Ann Heng, and Hui Huang.
\newblock Cascaded feature network for semantic segmentation of {RGB-D} images.
\newblock In {\em IEEE International Conference on Computer Vision (ICCV)},
  pages 1311--1319, 2017.

\bibitem{liu2020learning}
Nian Liu, Ni Zhang, and Junwei Han.
\newblock Learning selective self-mutual attention for {RGB-D} saliency
  detection.
\newblock In {\em IEEE Conference on Computer Vision and Pattern Recognition
  (CVPR)}, pages 13756--13765, 2020.

\bibitem{long2015fully}
Jonathan Long, Evan Shelhamer, and Trevor Darrell.
\newblock Fully convolutional networks for semantic segmentation.
\newblock In {\em IEEE Conference on Computer Vision and Pattern Recognition
  (CVPR)}, pages 3431--3440, 2015.

\bibitem{yunqiu_cod21}
Yunqiu Lv, Jing Zhang, Yuchao Dai, Aixuan Li, Bowen Liu, Nick Barnes, and
  Deng-Ping Fan.
\newblock Simultaneously localize, segment and rank the camouflaged objects.
\newblock In {\em IEEE Conference on Computer Vision and Pattern Recognition
  (CVPR)}, 2021.

\bibitem{mei2021Ming}
Haiyang Mei, Ge-Peng Ji, Ziqi Wei, Xin Yang, Xiaopeng Wei, and Deng-Ping Fan.
\newblock Camouflaged object segmentation with distraction mining.
\newblock In {\em IEEE Conference on Computer Vision and Pattern Recognition
  (CVPR)}, pages 8772--8781, 2021.

\bibitem{nafus2015hiding}
Melia~G Nafus, Jennifer~M Germano, Jeanette~A Perry, Brian~D Todd, Allyson
  Walsh, and Ronald~R Swaisgood.
\newblock Hiding in plain sight: a study on camouflage and habitat selection in
  a slow-moving desert herbivore.
\newblock {\em Behavioral Ecology}, 26(5):1389--1394, 2015.

\bibitem{niu2012leveraging}
Yuzhen Niu, Yujie Geng, Xueqing Li, and Feng Liu.
\newblock Leveraging stereopsis for saliency analysis.
\newblock In {\em IEEE Conference on Computer Vision and Pattern Recognition
  (CVPR)}, pages 454--461, 2012.

\bibitem{park2017rdfnet}
Seong-Jin Park, Ki-Sang Hong, and Seungyong Lee.
\newblock {RDFNet}: {RGB-D} multi-level residual feature fusion for indoor
  semantic segmentation.
\newblock In {\em IEEE International Conference on Computer Vision (ICCV)},
  pages 4980--4989, 2017.

\bibitem{penacchio2015three}
Olivier Penacchio, P~George Lovell, Innes~C Cuthill, Graeme~D Ruxton, and
  Julie~M Harris.
\newblock Three-dimensional camouflage: exploiting photons to conceal form.
\newblock {\em The American Naturalist}, 186(4):553--563, 2015.

\bibitem{peng2014rgbd}
Houwen Peng, Bing Li, Weihua Xiong, Weiming Hu, and Rongrong Ji.
\newblock {RGBD} salient object detection: A benchmark and algorithms.
\newblock In {\em European Conference on Computer Vision (ECCV)}, pages
  92--109, 2014.

\bibitem{dmra_iccv2019}
Yongri {Piao}, Wei {Ji}, Jingjing {Li}, Miao {Zhang}, and Huchuan {Lu}.
\newblock Depth-induced multi-scale recurrent attention network for saliency
  detection.
\newblock In {\em IEEE International Conference on Computer Vision (ICCV)},
  pages 7254--7263, 2019.

\bibitem{piao2019depth}
Yongri Piao, Wei Ji, Jingjing Li, Miao Zhang, and Huchuan Lu.
\newblock Depth-induced multi-scale recurrent attention network for saliency
  detection.
\newblock In {\em IEEE International Conference on Computer Vision (ICCV)},
  pages 7254--7263, 2019.

\bibitem{pike2018quantifying}
Thomas~W Pike.
\newblock Quantifying camouflage and conspicuousness using visual salience.
\newblock {\em Methods in Ecology and Evolution}, 9(8):1883--1895, 2018.

\bibitem{price2019background}
Natasha Price, Samuel Green, Jolyon Troscianko, Tom Tregenza, and Martin
  Stevens.
\newblock Background matching and disruptive coloration as habitat-specific
  strategies for camouflage.
\newblock {\em Scientific reports}, 9(1):1--10, 2019.

\bibitem{qian2020bi}
Chen Qian, Hongsheng Li, and Gang Zeng.
\newblock Bi-directional cross-modality feature propagation with
  separation-and-aggregation gate for {RGB-D} semantic segmentation.
\newblock In {\em European Conference on Computer Vision (ECCV)}, pages
  561--–577, 2020.

\bibitem{MiDaS_Ranftl_2020_TPAMI}
Ren\'{e} Ranftl, Katrin Lasinger, David Hafner, Konrad Schindler, and Vladlen
  Koltun.
\newblock Towards robust monocular depth estimation: Mixing datasets for
  zero-shot cross-dataset transfer.
\newblock {\em IEEE Transactions on Pattern Analysis and Machine Intelligence
  (TPAMI)}, 2020.

\bibitem{redmon2016you}
Joseph Redmon, Santosh Divvala, Ross Girshick, and Ali Farhadi.
\newblock You only look once: Unified, real-time object detection.
\newblock In {\em IEEE Conference on Computer Vision and Pattern Recognition
  (CVPR)}, pages 779--788, 2016.

\bibitem{ren2015exploiting}
Jianqiang Ren, Xiaojin Gong, Lu Yu, Wenhui Zhou, and Michael Ying~Yang.
\newblock Exploiting global priors for {RGB-D} saliency detection.
\newblock In {\em IEEE Conference on Computer Vision and Pattern Recognition
  (CVPR) Workshop}, pages 25--32, 2015.

\bibitem{ren2012rgb}
Xiaofeng Ren, Liefeng Bo, and Dieter Fox.
\newblock {RGB-(D)} scene labeling: Features and algorithms.
\newblock In {\em IEEE Conference on Computer Vision and Pattern Recognition
  (CVPR)}, pages 2759--2766, 2012.

\bibitem{ronneberger2015u}
Olaf Ronneberger, Philipp Fischer, and Thomas Brox.
\newblock {U-Net}: Convolutional networks for biomedical image segmentation.
\newblock In {\em International Conference on Medical image computing and
  computer-assisted intervention}, pages 234--241, 2015.

\bibitem{shen2021real}
Yichen Shen, Zhilu Zhang, Mert~R Sabuncu, and Lin Sun.
\newblock Real-time uncertainty estimation in computer vision via
  uncertainty-aware distribution distillation.
\newblock In {\em IEEE Winter Conference on Applications of Computer Vision
  (WACV)}, pages 707--716, 2021.

\bibitem{silberman2011indoor}
Nathan Silberman and Rob Fergus.
\newblock Indoor scene segmentation using a structured light sensor.
\newblock In {\em IEEE International Conference on Computer Vision (ICCV)
  Workshop}, pages 601--608, 2011.

\bibitem{Chameleon2018}
Przemysław Skurowski, Hassan Abdulameer, Jakub Baszczyk, Tomasz Depta, Adam
  Kornacki, and Przemysław Kozie.
\newblock Animal camouflage analysis: Chameleon database.
\newblock In {\em Unpublished Manuscript}, 2018.

\bibitem{tankus2001convexity}
Ariel Tankus and Yehezkel Yeshurun.
\newblock Convexity-based visual camouflage breaking.
\newblock {\em Computer Vision and Image Understanding (CVIU)}, 82(3):208--237,
  2001.

\bibitem{thayer1918concealing}
Gerald~Handerson Thayer.
\newblock {\em Concealing-coloration in the animal kingdom: an exposition of
  the laws of disguise through color and pattern}.
\newblock Macmillan Company, 1918.

\bibitem{valada2019self}
Abhinav Valada, Rohit Mohan, and Wolfram Burgard.
\newblock Self-supervised model adaptation for multimodal semantic
  segmentation.
\newblock {\em International Journal of Computer Vision (IJCV)}, pages 1--47,
  2019.

\bibitem{wang2017rgb}
Anzhi Wang and Minghui Wang.
\newblock {RGB-D} salient object detection via minimum barrier distance
  transform and saliency fusion.
\newblock {\em IEEE Signal Processing Letters (SPL)}, 24(5):663--667, 2017.

\bibitem{wei2020f3net}
Jun Wei, Shuhui Wang, and Qingming Huang.
\newblock {F$^3$Net}: Fusion, feedback and focus for salient object detection.
\newblock In {\em AAAI Conference on Artificial Intelligence (AAAI)}, pages
  12321--12328, 2020.

\bibitem{wilson2019experimental}
Evan~C Wilson, Amy~A Shipley, Benjamin Zuckerberg, M~Zachariah Peery, and
  Jonathan~N Pauli.
\newblock An experimental translocation identifies habitat features that buffer
  camouflage mismatch in snowshoe hares.
\newblock {\em Conservation Letters}, 12(2):12614, 2019.

\bibitem{xue2016camouflage}
Feng Xue, Chengxi Yong, Shan Xu, Hao Dong, Yuetong Luo, and Wei Jia.
\newblock Camouflage performance analysis and evaluation framework based on
  features fusion.
\newblock {\em Multimedia Tools and Applications}, 75(7):4065--4082, 2016.

\bibitem{yan2020mirrornet}
Jinnan Yan, Trung-Nghia Le, Khanh-Duy Nguyen, Minh-Triet Tran, Thanh-Toan Do,
  and Tam~V Nguyen.
\newblock Mirrornet: Bio-inspired adversarial attack for camouflaged object
  segmentation.
\newblock {\em arXiv preprint arXiv:2007.12881}, 2020.

\bibitem{Yang_2021_ICCV_COD_Uncertainty}
Fan Yang, Qiang Zhai, Xin Li, Rui Huang, Ao Luo, Hong Cheng, and Deng-Ping Fan.
\newblock Uncertainty-guided transformer reasoning for camouflaged object
  detection.
\newblock In {\em IEEE International Conference on Computer Vision (ICCV)},
  pages 4146--4155, 2021.

\bibitem{zhai2021Mutual}
Qiang Zhai, Xin Li, Fan Yang, Chenglizhao Chen, Hong Cheng, and Deng-Ping Fan.
\newblock Mutual graph learning for camouflaged object detection.
\newblock In {\em IEEE Conference on Computer Vision and Pattern Recognition
  (CVPR)}, pages 12997--13007, 2021.

\bibitem{ucnet_sal}
Jing Zhang, Deng-Ping Fan, Yuchao Dai, Saeed Anwar, Fatemeh~Sadat Saleh, Tong
  Zhang, and Nick Barnes.
\newblock {UC-Net}: Uncertainty inspired {RGB-D} saliency detection via
  conditional variational autoencoders.
\newblock In {\em IEEE Conference on Computer Vision and Pattern Recognition
  (CVPR)}, pages 8582--8591, 2020.

\bibitem{ssf_cvpr2020}
Miao Zhang, Weisong Ren, Yongri Piao, Zhengkun Rong, and Huchuan Lu.
\newblock Select, supplement and focus for {RGB-D} saliency detection.
\newblock In {\em IEEE Conference on Computer Vision and Pattern Recognition
  (CVPR)}, pages 3472--3481, 2020.

\bibitem{zheng2018detection}
Yunfei Zheng, Xiongwei Zhang, Feng Wang, Tieyong Cao, Meng Sun, and Xiaobing
  Wang.
\newblock Detection of people with camouflage pattern via dense deconvolution
  network.
\newblock {\em IEEE Signal Processing Letters (SPL)}, 26(1):29--33, 2018.

\end{thebibliography}
}

\end{document}